\documentclass[lettersize,journal,final]{IEEEtran}
\usepackage{amsmath,amsfonts}
\usepackage{amssymb}
\usepackage{amsthm}
\usepackage[ruled, lined, linesnumbered, commentsnumbered, longend]{algorithm2e}
\usepackage{array,multirow}
\usepackage{graphicx}
\usepackage[caption=false,font=small,labelfont=sf,textfont=sf]{subfig}
\usepackage[font=small,labelfont=bf]{caption}
\usepackage{textcomp}
\usepackage{stfloats}
\usepackage{url}
\usepackage{verbatim}
\usepackage{cite}
\usepackage{tabularx}
\usepackage{xcolor}
\usepackage{siunitx}
\usepackage{makecell}
\usepackage{comment}
\usepackage{float}
\usepackage{booktabs}
\usepackage{algpseudocode}
\usepackage{mathtools}
\usepackage{bm}
\usepackage{epsfig} 
\usepackage{times}
\usepackage{listings}
\usepackage{psfrag}
\usepackage[colorlinks,linkcolor=black,citecolor=black,urlcolor=black,bookmarks=false,hypertexnames=true]{hyperref} 
\usepackage{xparse}
\usepackage[english]{babel}
\usepackage{myMacros}
\usepackage{tab_metrics}

\newcolumntype{C}{>{$\displaystyle}c<{$}}
\makeatletter
\NewDocumentCommand{\LeftComment}{s m}{%
  \Statex \IfBooleanF{#1}{\hspace*{\ALG@thistlm}}\(\triangleright\) #2}
\makeatother

\newcommand{\bs}{\boldsymbol}

\newcommand{\jac}{\bs{\mathcal{J}}}

\newcommand{\argminD}{\arg\!\min}

\newtheorem{remark}{Remark}

\begin{document}
\title{GNSS/Multi-Sensor Fusion Using Continuous-Time Factor Graph Optimization for Robust Localization}

\author{Haoming Zhang$^{1}$,~\IEEEmembership{Member,~IEEE}, Chih-Chun Chen$^{1}$,~\IEEEmembership{Graduate Student Member,~IEEE},\\ 
Heike Vallery$^{1,2}$,~\IEEEmembership{Member,~IEEE}, and
Timothy~D.~Barfoot$^{3}$,~\IEEEmembership{Fellow,~IEEE}

\thanks{$^{1}$H. Zhang, C.-C. Chen, and H. Vallery are with the Institute of Automatic Control, Faculty of Mechanical Engineering, RWTH Aachen University, Aachen, Germany. (e-mail: \{c.chen, h.vallery\}@irt.rwth-aachen.de)}
\thanks{$^{2}$H. Vallery is also with the Department of BioMechanical Engineering, Delft University of Technology, and with the Department for Rehabilitation Medicine, Erasmus MC, Rotterdam, The Netherlands.}
\thanks{$^{3}$T. D. Barfoot is with the University of Toronto Robotics Institute, Toronto, Canada. (e-mail: tim.barfoot@utoronto.ca)}
}



\maketitle

\begin{abstract}
Accurate and robust vehicle localization in highly urbanized areas is challenging. Sensors are often corrupted in those complicated and large-scale environments. This paper introduces \texttt{gnssFGO}, a global and online trajectory estimator that fuses GNSS observations alongside multiple sensor measurements for robust vehicle localization. In \texttt{gnssFGO}, we fuse asynchronous sensor measurements into the graph with a continuous-time trajectory representation using Gaussian process regression. This enables querying states at arbitrary timestamps without strict state and measurement synchronization. Thus, the proposed method presents a generalized factor graph for multi-sensor fusion. To evaluate and study different GNSS fusion strategies, we fuse GNSS measurements in loose and tight coupling with a speed sensor, IMU, and lidar-odometry. We employed datasets from measurement campaigns in Aachen, Düsseldorf, and Cologne and presented comprehensive discussions on sensor observations, smoother types, and hyperparameter tuning. Our results show that the proposed approach enables robust trajectory estimation in dense urban areas where a classic multi-sensor fusion method fails due to sensor degradation. In a test sequence containing a $\SI{17}{\km}$ route through Aachen, the proposed method results in a mean 2-D positioning error $\SI{0.48}{\m}$ while fusing raw GNSS observations with lidar odometry in a tight coupling.
\end{abstract}

\begin{IEEEkeywords}
Sensor Fusion, Localization, Autonomous Vehicle Navigation, Factor Graph Optimization, GNSS
\end{IEEEkeywords}


\section{Introduction}
Safe and reliable autonomous driving operations in urban areas require accurate and consistent vehicle localization that infers a smooth trajectory estimate for planning and control tasks. Autonomous vehicles may use global navigation satellite systems (GNSS) to achieve global positioning in large-scale environments. However, the performance of GNSS is highly degraded when a vehicle passes through tunnels or urban canyons, where GNSS signal loss can be expected, greatly penalizing positioning availability. Moreover, the error dynamics of GNSS observations grow increasingly complex due to multipath and non-line-of-sight effects, resulting in inconsistent error models used in state estimation \cite{GNSS_Integrity_Performance}.
\begin{figure}[H]
\centering
\subfloat[Lidar map and trajectory using tightly coupled \texttt{gnssFGO} (ours). The proposed method provides robust trajectory estimation and clear lidar maps in GNSS-corrupted areas.]{\includegraphics[width=0.48\textwidth]{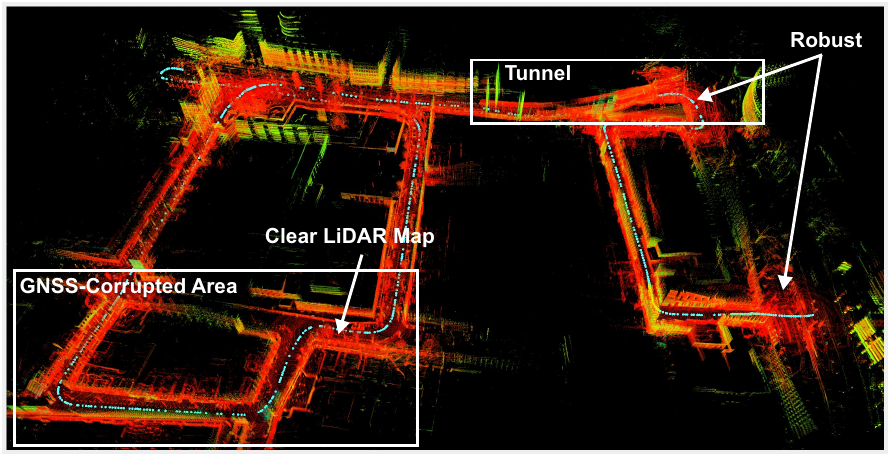}\label{fig: c02_gnssfgo_tc}}\hskip1ex
\subfloat[Lidar map and trajectory using LIO-SAM\cite{liosam}. This approach fused with GNSS positioning failed due to faulty scan registrations while crossing a tunnel and degrades dramatically with corrupted GNSS measurements.]{\includegraphics[width=0.48\textwidth]{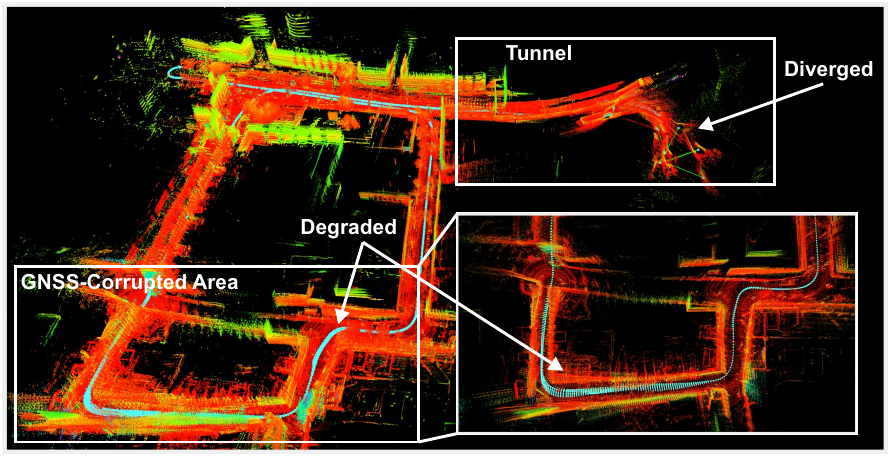}\label{fig: c02_liosam}}\hskip1ex
\subfloat[Measured and estimated trajectories. Fisheye images show challenging areas where GNSS observations are blocked or strongly corrupted.]{\includegraphics[width=0.48\textwidth]{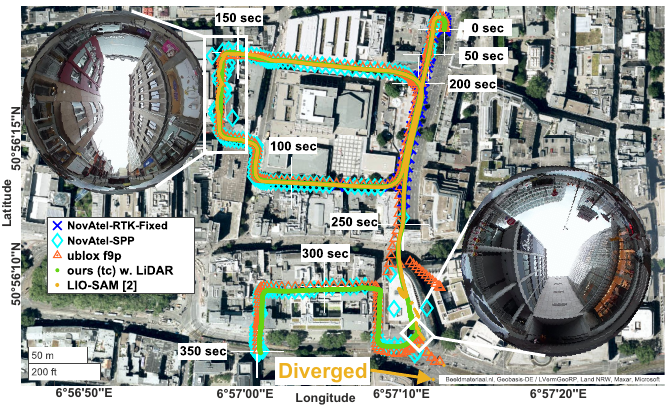}\label{fig:sub2}}\hskip1ex
\caption{Demonstration of multi-sensor fusion for vehicle localization of test sequence C02 in the city of Cologne.}
\label{fig: demo_liosam_tcfgo}
\end{figure}
 
Many previous works fuse information from local optical sensors (e.g., lidars or cameras) for vehicle localization. They can typically be categorized into pose retrieval using a given map \cite{map_based_local} and simultaneous location and mapping (SLAM) \cite{slam_vehicle}. Generally, landmarks in sensor frames are extracted and associated to acquire either frame-to-map global pose constraints or frame-to-frame local motion increments. Lacking high-quality maps for vehicle pose retrieval in many areas, approaches relying on local sensors can often only achieve satisfactory localization if the ground is even and sufficient loop-closure constraints help eliminate drift. However, these requirements cannot always be met for long-term autonomous operations in large-scale environments \cite{slam_largescale}.

In recent years, combining local sensors with GNSS has been investigated as a robust way to enable accurate and precise vehicle location in challenging areas. Incremental batch estimation implemented as factor graph optimization (FGO) is often superior to classic filtering-based algorithms in terms of localization performance and consistency \cite{ekf_fgo_compare, fgo_zhang}. Unlike Bayesian filters, a factor graph fuses prior information and sensor measurements associated with the to-be-estimated state variables into probabilistic representations. A maximum-a-posterior problem (MAP) can be formulated from the factor graph and solved in a batch configuration using iterative Gauss-Newton-like algorithms \cite{factor_graph_review}. In general, this optimization procedure is activated only if new sensor observations are available. Thus, previous works that use FGO generally rely on a primary sensor that schedules the optimization procedure \cite{liosam, hdl_slam, gvins, gnss_lidar_hsu, wen_vis_gnss}.

By this means, the primary sensor that is expected to effectively constrain the vehicle states coordinates the creation of new to-be-estimated state variables and initiates the solver for the MAP problem. To fuse additional sensor modalities, asynchronous measurements must be synchronized with the primary sensor, leading to information loss and inefficient fusion mechanisms. Furthermore, classic FGO approaches degrade if the primary sensor is compromised or fails, which is likely in challenging environments. In this case, state variables cannot be effectively constrained by other sensor observations if the graph is not constructed in time. Fig.\,\ref{fig: demo_liosam_tcfgo} exemplifies this problem, where a state-of-the-art lidar-centric SLAM approach diverges due to scan registration failures while driving in a tunnel\footnotemark[1]. In fact, as discussed in \cite{hsu_analysis_NLOS, keenan_iros22, urban_data_KAIST}, commonly used sensors in localization deteriorate under challenging environmental conditions, complicating robust and long-term vehicle localization.

\footnotetext[1]{Same noise models and smoother were used while benchmarking the lidar-centric approach with tightly coupled \texttt{gnssFGO}. The same illustration settings (e.g., point size) were used in Fig.\,\ref{fig: c02_liosam} and \ref{fig: c02_gnssfgo_tc}.}

In this work, we address the degradation problem of GNSS-based localization approaches by translating classic FGO for multi-sensor fusion into an approach where the graph associated with all to-be-estimated state variables is constructed deterministically based on a priori chosen timestamps. It thus presents a time-centric factor graph construction that is independent of any particular reference sensor (e.g., GNSS). To achieve this, we represent the vehicle trajectory in continuous time using a Gaussian process (GP). This approach incorporates a motion prior using the white-noise-on-jerk (WNOJ) motion model, as originally proposed in \cite{WNOJ}. The algorithm feeds new observations from each sensor independently into the factor graph without measurement-to-state synchronization. If a measurement cannot be temporally aligned with any state variable, we query a GP-interpolated state corresponding to the measurement used for the error evaluation.

To retrieve a robust global trajectory estimation while the GNSS measurements are strongly corrupted, we implemented the time-centric factor graph to fuse GNSS observations with measurements of an inertial measurement unit (IMU), optical speed sensor, and lidar for vehicle localization in challenging urban scenarios. We propose two factor graph structures for both loosely and tightly coupled fusion of GNSS observations alongside other local sensor measurements, demonstrating the flexibility of the proposed \texttt{gnssFGO}. For the graph that considers the GNSS positioning solution of a low-grade GNSS receiver in the loose coupling, we fuse the pre-integrated IMU measurements, 2-D velocity measurements, and lidar odometry. In tightly coupled fusion, we replace GNSS solution factors with GNSS pseudorange and Doppler-shift factors, which are expected to provide more effective constraints compared to inconsistent GNSS positioning in urban areas \cite{ekf_fgo_compare}. 

We used raw data from measurement campaigns in the cities of Aachen, Düsseldorf, and Cologne to evaluate the proposed approach by benchmarking with a well-known lidar-centric SLAM approach \cite{liosam, loam}. This lidar-centric SLAM has been shown to perform best for vehicle localization tasks in large-scale environments and can be configured to fuse GNSS measurements \cite{Geiger2012CVPR}, which presents an equivalent fusion mechanism as our loosely coupled \texttt{gnssFGO}. 

In contrast to our previous study \cite{fgo_zhang}, which focused only on trajectory smoothness using an offline FGO, we now address online multi-sensor fusion for vehicle localization. 

The contributions of this work are summarized as follows: 
\begin{enumerate}
    \item We propose a flexible, online, continuous-time factor graph optimization framework that can accommodate common multi-sensor fusion problems. The flexibility comes from the fact that (i) we can accommodate asynchronous measurements, and (ii) we choose estimation timestamps independent of any particular sensor frequency. This latter feature, as well as the smoothing effect of a motion prior, provides robustness in the presence of any particular sensor dropout.

    \item We implement the proposed method for vehicle localization in challenging scenarios and conduct comprehensive studies on loosely coupled and tightly coupled fusion mechanisms to fuse GNSS measurements with other local sensors to present extensive evaluations and discussions on accuracy, robustness, and run-time efficiency. 
    
    \item We evaluate the GP motion prior, which is implemented using the white-noise-on-acceleration (WNOA) and white-noise-on-jerk (WNOJ) models, to study the accuracy of the interpolated states.

    \item We introduce a scalable and modular estimation framework \texttt{gnssFGO}\footnotemark[2] that can be extended for arbitrary robot localization using continuous-time factor graph optimization.
\end{enumerate}

The rest of this article is organized as follows: Sec.\,\ref{sec: rw} presents a comprehensive literature review on multi-sensor fusion. Sec.\,\ref{sec: tc-fgo} introduces the proposed continuous-time FGO in detail. The mathematical background for factor formulations is presented in Sec.\,\ref{sec: metho}, whereas the graph implementations are introduced in Sec.\,\ref{sec: loca}. We verify our method in Sec.\,\ref{sec: exp} and conduct further experiments on the precision and consistency of estimated trajectories. Finally, Sec.\,\ref{sec: con} summarizes results and limitations. We release our code and raw data in our experiments\footnotemark[2]. A demonstration video is also available\footnotemark[3].
\footnotetext[2]{[online] \url{https://github.com/rwth-irt/gnssFGO}}
\footnotetext[3]{[online] \url{https://youtu.be/JhxJc1NFN7g}}


\section{Related Work} \label{sec: rw}
\subsection{Graph Optimization for GNSS-based Vehicle Localization}
In recent years, fusing GNSS observations using factor graph optimization for robust vehicle localization has drawn great attention. Compared with filtering-based approaches, FGO conducts batch optimization, where all measurement models are re-linearized and re-evaluated iteratively, resulting in a more robust state estimation even with measurement outliers. Previous work demonstrated robust localization in urban areas only by factoring pseudoranges with robust error models \cite{gnss_fgo_basic1, gnss_fgo_basic2}. Later, Wen et al.,\cite{ekf_fgo_compare} and Zhang et al.,\cite{fgo_zhang} showed that FGO generally outperforms Kalman filters with respect to the precision and smoothness of the estimated trajectory. 

GNSS data can be integrated into the graph using a loosely or tightly coupled schema \cite{groves}. While the loosely coupled fusion incorporates GNSS positioning solution into the graph, pre-processed raw GNSS observations such as code or carrier-phase measurements can be fed into the estimator in a tight coupling as state constraints. As the to-be-estimated state variables can be directly observed in GNSS solutions, fusing GNSS data in a loose coupling enables quick convergence and elevated accuracy if high-quality Real-Time-Kinematic (RTK)-fixed GNSS solutions are available. 

In contrast, the integration of raw GNSS observations contributes to multiple state constraints associated with received satellites, which has been shown to be more robust than loose coupling \cite{lc_tc_performance, ekf_fgo_compare, weisong_2019}. Wen et al.\,\cite{hsu_rtk_fgo} included double-differenced pseudorange (DDPR) and double-differenced carrier-phase measurements (DDCP) in FGO, resulting in performance improvement. Later, this work was extended to efficiently model carrier-phase constraints between multiple satellite measurement epochs within a time window \cite{hsu_rtk2_fgo}. In \cite{taro_gnss_odom}, time-differenced carrier-phase (TDCP) was integrated with the cycle-slip estimation, which achieved accurate localization while presenting substantial availability compared to DDCP if satellites can be continuously tracked. Congram and Barfoot \cite{benjamin_gps_odom} also proposed a global positioning system (GPS) odometry using TDCP with more prominent cycle slip detection and showed an effective drift reduction compared to visual odometry. However, since carrier-phase observations are also disturbed in deep urban areas, the robustness of the state estimation cannot yet be guaranteed.  

As factor graph optimization presents a convenient tool for robust error modeling \cite{fgo_robust}, several works employ m-Estimators to reject faulty GNSS observations \cite{gnss_fgo_basic2, taro_gnss_odom, taro_1st, fgo_zhang}. Recently, FGO has been explored in the context of vehicle location based on GNSS for noise distribution identification or adaptive rejection of outliers \cite{gmm_incr, fgo_gnc}, showing a positive impact on consistent trajectory estimation using FGO.

\subsection{Graph Optimization for Multi-Sensor Fusion}
While the aforementioned works have particularly explored graph optimization for GNSS observations, they may still suffer from performance degeneration in complex scenarios if GNSS measurements are lost or present outliers. Therefore, another research domain focuses on fusing more sensor modalities (more than two) alongside GNSS observations into the graph, with applications predominantly in SLAM.

A pose graph that fuses GPS position measurements and lidar odometry with loop-closure constraints for outdoor scenarios improved both runtime efficiency and performance compared to lidar-only approaches \cite{hdl_slam}. In \cite{liosam}, feature-based lidar odometry and loop-closure constraints were merged into a factor graph with synchronized GPS position measurements to achieve a drift-free pose estimate, which was forwarded to another graph optimization with pre-integrated IMU measurements for high-frequency and real-time state estimation. In addition to integrating feature-based lidar odometry into FGO, the lidar map can also be used for GNSS visibility assessment \cite{gnss_lidar_hsu}. Some works also introduce camera-centric sensor fusion, where other sensor observations are synchronized with camera data and fused on the graph \cite{gnss_vio_adaptive, gvins, wen_vis_gnss}. In \cite{multisensor1}, camera, lidar, and wheel odometers were fused into the graph along with the GNSS positioning solution and IMU measurements, presenting consistent localization in featureless environments for long-term runs. Similar works also conduct multi-sensor fusion without GNSS and propose a carefully managed fusion architecture \cite{lidar_vio_inertial_any, vio_lidar_withorder}. However, these works still require well-handled data synchronization and careful graph construction to fuse heterogeneous sensor measurements.

Many recent approaches introduce multi-graph structures to achieve flexible and compact sensor fusion. In \cite{multisensor_construction}, IMU, GNSS, and lidar observations were separately integrated into multiple graphs in parallel with a switching mechanism. When the GNSS receiver lost its signal, the lidar-centric graph was activated. Another work aimed to confederate loosely and tightly coupled fusion schemes to ensure estimation performance \cite{superodom}. Each sensor modality is associated with a separate graph and proposes odometry factors to the IMU-centric graph that provides final estimated states in real-time. However, incorporating multi-graph structures introduces redundant and complex system architectures that may require well-managed engineering work. Moreover, these works did not address challenging environments where sensor observations can be highly corrupted with inconsistent noise distributions.

Other works exploit high-frequency IMU measurements to coordinate multi-sensor fusion. In \cite{VI_GPS_project} and \cite{VI_GPS_dropout}, asynchronous global pose measurements (e.g., GNSS measurements) are propagated into timestamps of visual-inertial factors using pre-integrated IMU measurements. The same concept has been extended to a forward-backward IMU pre-integration mechanism in order to precisely associate asynchronous measurements with keyframes \cite{multisensor_propagation}. Nevertheless, these methods still depend on the noisy IMU sensor, which introduces uncertainty. 

\subsection{Continuous-Time Trajectory Representation}
One essential requirement for flexible graph-based multi-sensor fusion is the ability to query the states associated with the observations within the iterative optimization process. This requirement can be fulfilled if the trajectory is represented in continuous time. In \cite{continuous_time_basis_func}, B-splines were proposed as a parametric approach to represent the trajectory in continuous time. This method was later used to propose stereo-inertial odometry \cite{conti_vslam}. Another approach utilizes exactly sparse Gaussian process (GP) regression by assuming that system dynamics follows a linear time-varying stochastic differential equation (LTV-SDE) \cite{Barfoot2014BatchCT}. The system dynamics is typically modeled as white-noise-on-acceleration (WNOA). This approach was verified in \cite{STEAP, gp-dong, Dong20174DCM}, where the reliability of this proposed surrogate dynamics model was demonstrated. Recently, Tang et al.\,\cite{WNOJ} proposed an improved system dynamics model, which assumed a white-noise-on-jerk (WNOJ) model in LTV-SDE. They showed that the WNOJ could model the vehicle dynamics more accurately and thus was appropriate for systems with more complicated dynamics. As discussed in \cite{cont_traj_compare}, continuous-time trajectory using GPs should be used if the measurement times match the estimation times. Thus, we follow this aspect and adapt the GP-WNOJ model proposed in \cite{WNOJ} as between-state motion constraints and state interpolator to fuse asynchronous measurements. Although continuous-time trajectory representation is studied for localization and mapping problems by extending incremental smoothing using sparse GP interpolation to reduce computation time \cite{gp-dong}, fusing GNSS observations with multiple heterogeneous sensor measurements for online vehicle localization has not yet been presented or discussed. 

\subsection{Modular Estimation Framework}
As the aforementioned approaches share similar procedures to solve estimation problems, the idea of a modular estimation framework that unifies the system design for different applications has emerged. In \cite{rtabmap}, Labbé and Michaud originally proposed a real-time mapping framework to manage the memory of the internal map for loop closure detection. This framework has been continuously extended by the same authors in \cite{rtabmap2} to enable multi-sensor fusion and benchmarking. Due to its modularity and scalability, many works based on this framework can be performed \cite{rtabmap_derive1, rtabmap_derive2}. A similar framework for a plug-and-play SLAM system has been presented in \cite{pap_slam}. Recently, Solà et al. \cite{wolf} inherited this design with a tree-based estimation framework, which formulates all the necessary robot entities in different branches, including hardware, trajectory, and map management. Sensor measurements and prior information are fused using a decentralized strategy in which primary sensors are selected in the configuration file to actively create new keyframes (aka state variables). However, none of the modular frameworks mentioned above represents the trajectory in continuous time, which still requires measurement alignment to keyframes, making a loss of sensor observations inescapable.

Inspired by above mentioned works, we address the problem of multi-sensor fusion for GNSS-based vehicle localization using continuous-time trajectory representation, which enables a fusion of asynchronous sensor observations in a single factor graph. Our hypotheses of the contributions presented above are: (i) factor graph construction in continuous time generalizes multi-sensor fusion and enables consistent trajectory estimation that incorporates effective state constraints from multiple sensor modalities in challenging scenarios; (ii) in this spirit, a natural and efficient modular estimation framework can be presented thanks to the ability of querying state at arbitrary times, in which different applications and experiments can be configured directly using configuration files; (iii) the GP-WNOJ motion model presents a larger capacity to represent complicated system dynamics, such as driving in urban areas.

\section{Time-Centric Factor Graph Optimization} \label{sec: tc-fgo}
\begin{figure}[!t]
    \centering
    \includegraphics[width=0.45\textwidth]{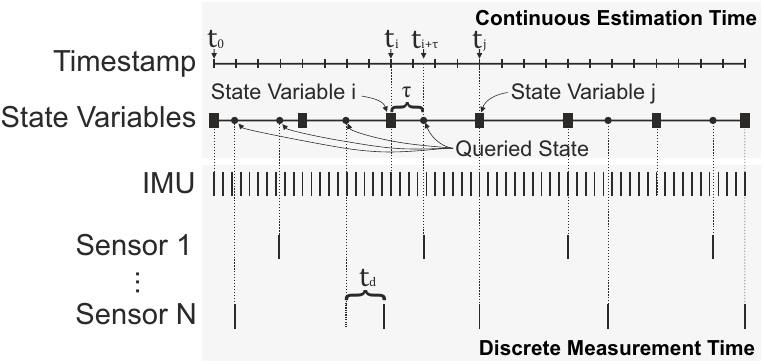}
    \caption{Continuous-time state estimation with asynchronous measurements. A time offset $\tau$ can be calculated with respect to a former state variable $\myFrameVec{x}{i}{}$ at timestamp $t_i$ for each asynchronous measurement. The variable $t_d$ denotes a measurement delay that is assumed to be given.}
    \label{fig: state_measu}
\end{figure}

In this section, we introduce an implementation of continuous-time trajectory estimation, as proposed in \cite{Barfoot2014BatchCT, WNOJ}. Generally, fusing multiple heterogeneous sensor observations into a state estimator incorporates different timestamps due to asynchronous measurements and unpredictable delays. In this work, we assume that the state estimator and all measurements have the same clock. Compared to the estimated states in continuous time, all sensor observations are sampled and processed in asynchronous timestamps, as illustrated in Fig.\,\ref{fig: state_measu}. 
Here, we use the variable $\tau = \myFrameScalar{t}{\rm meas.}{} - \myFrameScalar{t}{\myFrameVec{x}{i}{}}{}$ to define the time offset between a non-delayed measurement and the last state variable $\myFrameVec{x}{i}{}$ prior to this measurement. If a measurement is delayed with the timestamp $\check{t}_{\rm meas.}$, we use the given time delay $t_d$ to calculate the non-delayed measurement timestamp $\myFrameScalar{t}{\rm meas.}{}=\check{t}_{\rm meas.} - t_d$.

\begin{figure}[!t]
    \centering
    \includegraphics[width=0.45\textwidth]{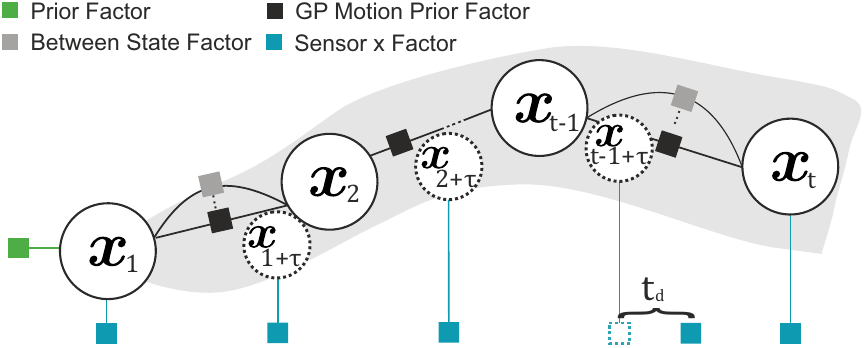}
    \caption{A general time-centric factor graph. The state variables $\myFrameVec{x}{t}{}$ are created and constrained with GP motion prior factors on time while all asynchronous measurements are fused by querying a state with a time offset $\tau$ between the measurement and the former state variable. The queried states (in dashed circles) are thus not to-be-estimated state variables. We assume that the measurement delay $t_d$ is known to correct the measurement timestamp for querying a state.}
    \label{fig: tcfgo}
\end{figure}

We employ GP motion priors that enable a continuous-time trajectory representation. In this way, constructing a factor graph can be deterministic and time-centric, bypassing asynchronous sensor frequencies and timing issues. We show the general structure of a time-centric factor graph in Fig.\,\ref{fig: tcfgo}, where the to-be-estimated state variables $\myFrameVec{x}{t}{}$ are presented in solid line circles on a continuous-time trajectory. Queried states in dashed line circles are not to-be-estimated state variables and, therefore, are only queried between two successive state variables using the time offset $\tau$ between the sensor timestamp and the previous state variable.

\renewcommand{\algorithmiccomment}[1]{/* \textit{#1} */} 
\SetKwInOut{KwIn}{Input}
\SetKwInOut{KwOut}{Output}

\begin{algorithm}[!t] 
\SetKwFunction{DROPPED}{DROPPED}
\SetKwFunction{CACHED}{CACHED}
\SetKwFunction{Synchronized}{SYNCHRONIZED}
\SetKwFunction{Interpolated}{INTERPOLATED}
\SetKwFunction{preinte}{doIMUPreIntegration}
\SetKwFunction{newstate}{NewStateVariable}
\SetKwFunction{DROPPED}{DROPPED}
\SetKwFunction{imufactor}{IMUFactor}
\SetKwFunction{sensorfactor}{SensorFactor}
\SetKwFunction{GPsensorfactor}{GPSensorFactor}
\SetKwFunction{between}{GPMotionFactor}
\SetKwFunction{init}{initGraph}
\SetKwFunction{dropObs}{discardMeasurement}
\SetKwFunction{cacheObs}{cacheMeasurement}
\SetKwFunction{initimu}{initIMUPreIntegrator}
\SetKwFunction{queryIndex}{queryStateInfo}
\SetKwFunction{doOpt}{doOptimizationAndMarginalization}
\SetKwFunction{getTime}{updateTimestamp}
\KwIn{Last state id and timestamp pair $(\myFrameScalar{x}{\mathrm{id}}{-},~\myFrameScalar{x}{\mathrm{ts}}{-})$ \\
      Propagated states $\myFrameVec{x}{k}{-}\in\bm{\mathcal{X}^{-}},~ k = 1...n$ \\
      List of sensor measurements $\myFrameScalar{s}{}{}\in\bm{\mathcal{S}}$}
\KwOut{Current state id and timestamp pair $(\myFrameScalar{x}{\mathrm{id}}{+},~\myFrameScalar{x}{\mathrm{ts}}{+})$ \\
       Optimized state $\myFrameVec{x}{k}{+}$ and uncertainties $\myFrameVec{P}{k}{+}$}
\caption{Time-centric factor graph optimization} \label{alg: TC-FGO}
$\bm{\mathcal{G}} \leftarrow \init(\myFrameVec{x}{0}{-},~\myFrameVec{P}{0}{-})$\;
List of state id and timestamp pairs $\bm{\mathcal{P}} = \emptyset$\;
$\myFrameScalar{x}{\mathrm{id}}{} = \myFrameScalar{x}{\mathrm{id}}{-}$\;
\For{$k = 1 : n$}  {
    $\myFrameScalar{x}{\mathrm{id}}{+} = \myFrameScalar{x}{\mathrm{id}}{} + 1$\;
    $\myFrameScalar{x}{\mathrm{ts}}{+} = \getTime(\myFrameScalar{x}{\mathrm{ts}}{-})$\;
    $\bm{\mathcal{G}} \leftarrow \newstate(\myFrameScalar{x}{\mathrm{id}}{},~\myFrameScalar{x}{\mathrm{ts}}{},~\bm{\mathcal{X}})$\;
    $\bm{\mathcal{G}} \leftarrow \between(\myFrameScalar{x}{\mathrm{id}-1}{},~\myFrameScalar{x}{\mathrm{id}}{})$\;
    $\bm{\mathcal{P}} \leftarrow (\myFrameScalar{x}{\mathrm{id}}{},~\myFrameScalar{x}{\mathrm{ts}}{})$\;
}
\For{Each sensor $\myFrameScalar{s}{}{} \in \bm{\mathcal{S}}$}{
    \For{Each observation $ k = 1 : m$}{
        $(\myFrameScalar{x}{i}{\mathrm{id}},~\myFrameScalar{x}{i}{\mathrm{ts}},~\tau,~type)$\
        \hspace*{1em} $\leftarrow \queryIndex(\myFrameScalar{\mathrm{timestamp}}{k}{\myFrameVec{s}{}{}},~\bm{\mathcal{P}})$\;
        \uIf{$type$ is $\DROPPED$}{
            \Comment{Measurements in the past.}\;
            $\dropObs(\myFrameVec{o}{k}{s})$\;
        }
        \uElseIf{$type$ is $\Synchronized$}{
            $\bm{\mathcal{G}} \leftarrow \sensorfactor(\myFrameScalar{x}{\mathrm{id}}{},~\myFrameVec{o}{k}{s})$\;
        }
        \uElseIf{$type$ is $\Interpolated$}{
            $\bm{\mathcal{G}} \leftarrow \GPsensorfactor(\myFrameScalar{x}{i}{\mathrm{id}},~\myFrameScalar{x}{i+1}{\mathrm{id}},~\tau,~\myFrameVec{o}{k}{\myFrameVec{s}{}{}})$\;
        }
        \uElseIf{$type$ is $\CACHED$}{ 
            \Comment{Measurements in the future.}\
            $\cacheObs(\myFrameVec{o}{k}{s})$\;
        }
    }
}
$ \{ \myFrameVec{x}{k}{+}, ~\myFrameVec{P}{k}{+} \} \leftarrow$\
$\hspace*{1em} \doOpt(\bm{\mathcal{G}})$\;
\Return{$ \{(\myFrameScalar{x}{\mathrm{id}}{+},~\myFrameScalar{x}{\mathrm{ts}}{+}),~ \myFrameVec{x}{k}{+},~\myFrameVec{P}{k}{+}\}$}
\end{algorithm}

Alg.\,\ref{alg: TC-FGO} explains one optimization procedure from graph construction to iterative optimization. Assume that the time-centric factor graph is extended with $n$ new to-be-estimated state variables in each procedure. We extend the graph with $n$ state variables and create GP motion prior factors that constrain the relative state transitions between two successive state variables. In doing so, the timestamps of all state variables are chosen deterministically. While solving the iterative optimization problem, an initial prediction $\myFrameVec{x}{k}{-}\in\bm{\mathcal{X}^{-}}$ must be provided for each state variable. These predictions can be acquired using prior motion models (e.g., GP state extrapolation \cite{vtr_gps_failure}). In this work, we utilize state propagation using IMU measurements to calculate the initial estimate of future states at high frequency.

As new sensor observations are received at different timestamps in parallel to estimation times, we retrieve the cached $m$ observations from each sensor $s \in \mathcal{S}$ in a second loop. We define a time threshold $\myFrameScalar{t}{\mathrm{sync}}{}$ for state-observation alignment to query the index of related state variables. If state variables can be associated with sensor observations within this threshold, normal sensor factors are added to the graph. Otherwise, we construct the measurement factors by querying a GP interpolated state aligned with the measurement timestamp. In this case, two successive state variables $\myFrameVec{x}{i}{}$ and $\myFrameVec{x}{j}{},~j = i+1$ are obtained with a time offset $\tau$ between the measurement and the former state $\myFrameVec{x}{i}{}$.

After graph construction, we employ a Gauss-Newton-like optimizer to solve the MAP problem \cite{fgo_kaess_book}. The optimized state $\myFrameVec{x}{k}{+}$ and marginalized uncertainties $\myFrameVec{P}{k}{+}$ are returned for further state propagation, as introduced in Sec.\,\ref{sec: sys_overview}.


\section{Mathematical Background} \label{sec: metho}

\subsection{Frame and Frame Transformation} \label{sec: frames}
Starting with GPS in 1987, GNSS typically provides a vehicle's position in the World Geodetic System (WGS84) frame using geodetic ellipsoidal (aka geodetic) coordinates (Latitude $\varphi$, Longitude $\lambda$, and Height $h$, LLH) \cite{gnss_data}, as shown in Fig.\,\ref{fig: frame}. Geodetic coordinates can be transformed into the Earth-centered, Earth-fixed (ECEF) frame that is defined at the center of Earth's mass, denoted as $(\cdot)^e$. As many estimation and control approaches require a Cartesian frame in a local tangent plane, the North-East-Down (NED) frame and the East-North-Up (ENU) frame are commonly introduced as navigation frames $(\cdot)^n$ to present the vehicle's velocity and orientation. In this work, we present the pose and velocity of the vehicle in the ECEF frame. A transformation from frame $e$ to frame $n$ is used to formulate factors and calculate error metrics.
We also introduce a local-world frame $(\cdot)^w$ following \cite{gvins} that rotates by the initial yaw angle of the vehicle in frame $n$. The body frame is denoted as $(\cdot)^b$. In the following, we briefly introduce the related frame transformations.

\subsubsection{Transform between Geodetic and Cartesian Coordinates}
A geodetic coordinate $\myFrameVec{p}{}{\rm LLH}=[\varphi~\lambda~h]^T$ can be transformed to the ECEF frame using \cite{groves}
\begin{align}
    \myFrameScalar{x}{b}{e} &= (\myFrameScalar{R}{e}{} + h)\cos{\varphi}\cos{\lambda} \label{eq: llh2ecef1}, \\
    \myFrameScalar{y}{b}{e} &= (\myFrameScalar{R}{e}{} + h)\cos{\varphi}\sin{\lambda} \label{eq: llh2ecef2}, \\
    \myFrameScalar{z}{b}{e} &= [\myFrameScalar{R}{e}{}(1-\mathrm{ecc}^2)+h]\sin{\varphi} \label{eq: llh2ecef3},
\end{align}
where the constant $\mathrm{ecc}=0.08181919$ is the eccentricity\footnotemark[4] of the ellipsoid. The transverse radius of curvature given the latitude $\varphi$ is calculated as $\myFrameScalar{R}{e}{}(\varphi) = \frac{a}{\sqrt{1-\mathrm{ecc}^2\sin^2(\varphi)}}$, where the scalar $a=\SI{6378137}{m}$ denotes the equatorial radius\footnotemark[4] of the Earth.

However, transforming a Cartesian coordinate in frame $e$ back to LLH using the inverse of \eqref{eq: llh2ecef1}-\eqref{eq: llh2ecef3} can only be solved iteratively due to the nonlinearity \cite{groves}. Although many closed-form alternatives for this transformation are available, we use Heikkinen's solution \cite{heikkinen} in this work considering its high precision \cite{ecef_llh_comparison}. Due to limited space, we refer the reader to \cite{heikkinen} or our code\footnotemark [2] for more details on the implementation.

\footnotetext[4]{The eccentricity and the Earth's equatorial radius in different satellite systems vary slightly depending on the ellipsoid and the satellite geometry. In this work, we follow the parameter definitions from WGS84 \cite{wgs84}.}

\subsubsection{Transform between Frame $e$ and Frame $n$}
\begin{figure}[!t]
    \centering
    \includegraphics[width=0.42\textwidth]{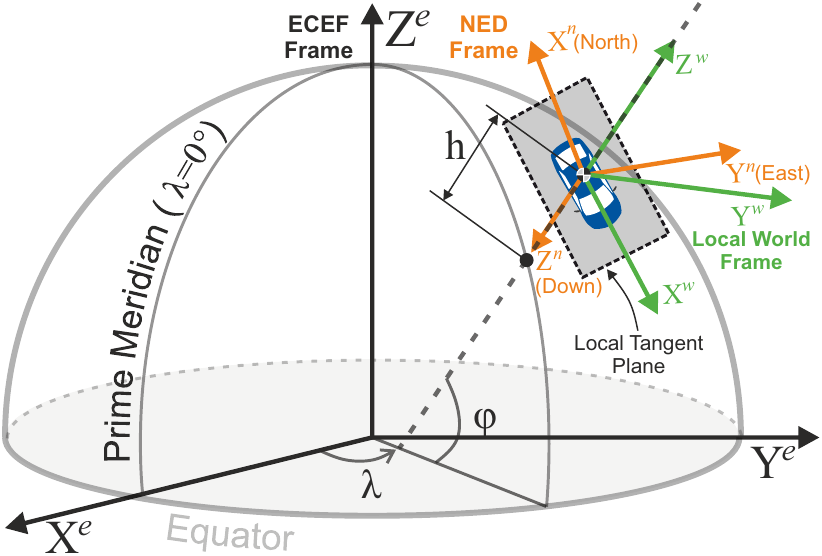}
    \caption{Coordinate frames used in this work.}
    \label{fig: frame}
\end{figure}

Given a point in the $e$ frame $\myFrameVec{p}{0}{e}$ as origin, a fixed navigation frame (aka local tangent plane) can be determined by two plane rotations associated with longitude $\lambda_0$ and latitude $\varphi_0$ of $\myFrameVec{p}{0}{\rm LLH}$. Fig.\,\ref{fig: frame} illustrates this transformation from the frame $e$ to the NED frame using the direction cosine matrix (DCM)
\begin{align}
\small
\myFrameVec{R}{e}{\rm ned}(\myFrameVec{p}{0}{\rm LLH}) = \left[
        \begin{matrix}
           -\sin{\varphi_0}\cos{\lambda_0}& -\sin{\varphi_0}\sin{\lambda_0} & \cos{\varphi_0}\\
           -\sin{\lambda_0} & \cos{\lambda_0}  & 0\\
           -\cos{\varphi_0}\cos{\varphi_0} & -\cos{\varphi_0}\sin{\lambda_0} & -\sin{\varphi_0} 
        \end{matrix}
    \right].
    \label{eq: R_e_n}
\end{align}
The DCM from frame e to the ENU frame is given as
\begin{align}
    \myFrameVec{R}{e}{\rm enu}(\myFrameVec{p}{0}{\rm LLH}) = \left[
        \begin{matrix}
           -\sin{\lambda_0} & \cos{\lambda_0}  & 0\\
           -\cos{\lambda_0}\sin{\varphi_0} & -\sin{\lambda_0}\sin{\varphi_0} & \cos{\varphi_0} \\
           \cos{\lambda_0}\cos{\varphi_0} & \sin{\lambda_0}\cos{\varphi_0} &  \sin{\varphi_0}
        \end{matrix}
    \right].
\end{align}

\subsection{Notation}
To present the state variables in different frames, we use $\myFrameVec{R}{b}{e} \in \mathbb{R}^{3\times3}$ and $\myFrameVec{p}{b}{e} \in \mathbb{R}^{3}$ to denote the rotation matrix and position vector of frame $b$ relative to frame $e$. This notation is extended to $\myFrameVec{R}{b,t}{e}$ to represent the states with respect to time $t$. For motion increments in the same frame, we simplify the notation as $\Delta\myFrameVec{p}{ij}{}$ to represent the translational offset of two timestamps $i$ and $j$. We follow the pose representation $\myFrameVec{T}{b}{e} = \Big[\begin{smallmatrix} \myFrameVec{R}{b}{e}&\myFrameVec{p}{b}{e}\\ \myFrameVec{0}{}{}&1 \end{smallmatrix}\Big] \in SE(3)$ to calculate the motion increment \cite{BarfootBook2024}. For high-dimensional transition matrices in GP motion models (e.g., $\myFrameVec{\Lambda}{}{}(\tau) \in \mathbb{R}^{18\times18}$), we denote the subblocks $\myFrameVec{\Lambda}{mn}{} \in \mathbb{R}^{6\times6}$ associated with different state components for linear state querying in \eqref{eq: gp_query}.

\subsection{Continuous-Time Trajectory Representation using GP} \label{sec: continuous-time-traj}
Barfoot et al. \cite{Barfoot2014BatchCT, fullsteam} originally proposed a continuous-time trajectory representation using Gaussian process regression, which presents an exactly sparse kernel by assuming the system dynamics follow a linear time-invariant stochastic differential equation (LTI-SDE):
\begin{align}
    \label{eq: ltvsde}
    \begin{split}
        \myFrameVecDot{\gamma}{}{}(t) &= \myFrameVec{A}{}{}\myFrameVec{\gamma}{}{}(t) + \myFrameVec{B}{}{}\myFrameVec{u}{}{}(t) + \myFrameVec{F}{}{}\myFrameVec{w}{}{}(t), \\
        \myFrameVec{w}{}{}(t) &\sim \mathcal{GP}(\myFrameVec{0}{}{},~\myFrameVec{Q}{c}{}\cdot\delta(t-t^{'})),
    \end{split}
\end{align}
where the vector $\myFrameVec{\gamma}{}{}(t)$ represents a local state variable. The time-varying system matrices are denoted as $\myFrameVec{A}{}{},~ \myFrameVec{B}{}{}$ and $\myFrameVec{F}{}{}$, respectively. The input vector $\myFrameVec{u}{}{}(t)$ is set to $\myFrameVec{0}{}{}$. The process noise $\myFrameVec{w}{}{}(t)$ is given as a zero-mean Gaussian process (GP) with the kernel function formulated with the power spectral density matrix $\myFrameVec{Q}{c}{} \in \mathbb{R}^{6\times6}$ and the Dirac delta function, $\delta$ \cite{BarfootBook2024}. 

In discrete time following \cite{fullsteam}, this state-space model can be furthermore interpreted to interpolate an arbitrary state $\myFrameVec{\gamma}{}{}(t_{i+\tau})$ at timestamp $t_{i+\tau} = t_i + \tau$ between two successive local states $\myFrameVec{\gamma}{}{}(t_i)$ and $\myFrameVec{\gamma}{}{}(t_j)$, where the state timestamps $t_i < t_{i+\tau} < t_j$, using
\begin{align}
    \myFrameVec{\gamma}{}{}(t_{i+\tau}) = \myFrameVec{\Lambda}{}{}(t_{i+\tau}) \myFrameVec{\gamma}{}{}(t_i) + \myFrameVec{\Omega}{}{}(t_{i+\tau})\myFrameVec{\gamma}{}{}(t_{j}),
    \label{eq:gp_discrete}
\end{align}
where
\begin{align}
    \label{eq:gp_lambda_psi}
    \myFrameVec{\Lambda}{}{}(t_{i+\tau}) &= \myFrameVec{\Phi}{}{}(\tau) - \myFrameVec{\Omega}{}{}(t_{i+\tau})\myFrameVec{\Phi}{}{}(t_j-t_{i+\tau}), \\
    \myFrameVec{\Omega}{}{}(t_{i+\tau}) &= \myFrameVec{Q}{}{}(\tau)\myFrameVec{\Phi}{}{}(t_j-t_{i+\tau})^T\myFrameVec{Q}{}{-1}(\tau).\label{eq:gp_Phi}
\end{align}
The system transition matrix $\myFrameVec{\Phi}{}{}$ in \eqref{eq:gp_lambda_psi} and \eqref{eq:gp_Phi} can be defined using a white-noise-on-acceleration (WNOA, aka constant-velocity) prior, as demonstrated in earlier works \cite{Barfoot2014BatchCT, fullsteam}. Later, Tang et al. \cite{WNOJ} introduced a white-noise-on-jerk (WNOJ) prior that presents third-order system dynamics with the system transition function
\begin{align}
    \label{eq:gp_phi}
    \myFrameVec{\Phi}{}{}(\Delta t) =  \left[\begin{matrix}
                                            \myFrameVec{1}{}{} & \Delta t\myFrameVec{1}{}{} & \frac{1}{2}\Delta t^2\myFrameVec{1}{}{}\\
                                            \myFrameVec{0}{}{} & \myFrameVec{1}{}{}  & \Delta t\myFrameVec{1}{}{} \\
                                            \myFrameVec{0}{}{} & \myFrameVec{0}{}{} & \myFrameVec{1}{}{} 
                                            \end{matrix} \right].
\end{align} 
The time-varying covariance matrix $\myFrameVec{Q}{}{}(\Delta t) \in \mathbb{R}^{18\times18}$ and its precision matrix $\myFrameVec{Q}{}{-1}(\Delta t)$ are computed as
\begin{align}
    \label{eq:gp_Q}
    \myFrameVec{Q}{}{}(\Delta t) = \left[ \begin{matrix} 
    \frac{1}{20}\Delta t^5\myFrameVec{Q}{c}{} & \frac{1}{8}\Delta t^4\myFrameVec{Q}{c}{} & \frac{1}{6}\Delta t^3\myFrameVec{Q}{c}{}\\
    \frac{1}{8}\Delta t^4\myFrameVec{Q}{c}{}  & \frac{1}{3}\Delta t^3\myFrameVec{Q}{c}{} & \frac{1}{2}\Delta t^2\myFrameVec{Q}{c}{}\\
    \frac{1}{6}\Delta t^3\myFrameVec{Q}{c}{}  & \frac{1}{2}\Delta t^2\myFrameVec{Q}{c}{} & \Delta t\myFrameVec{Q}{c}{}
    \end{matrix}  \right],
\end{align}
\begin{footnotesize}
\begin{align}
    \myFrameVec{Q}{}{-1}(\Delta t) = \left[ \begin{matrix} 
                           720\Delta t^{-5}\myFrameVec{Q}{c}{-1} & -360\Delta t^{-4}\myFrameVec{Q}{c}{-1} & 60\Delta t^{-3}\myFrameVec{Q}{c}{-1}\\[2pt]
                           -360\Delta t^{-4}\myFrameVec{Q}{c}{-1}  & 192\Delta t^{-3}\myFrameVec{Q}{c}{-1} & -36\Delta t^{-2}\myFrameVec{Q}{c}{-1}\\[2pt]
                           60\Delta t^{-3}\myFrameVec{Q}{c}{-1}  & -36\Delta t^{-2}\myFrameVec{Q}{c}{-1} & 9\Delta t^{-1}\myFrameVec{Q}{c}{-1}
                           \end{matrix}  \right].
\end{align}
\end{footnotesize}
Compared to other approaches, trajectory representation (interpolation) using Gaussian process regression effectively incorporates physics-driven models to retrieve realistic vehicle motion by scaling the transition function with the time-varying covariance matrix $\myFrameVec{Q}{}{}$. As the hyper-parameter $\myFrameVec{Q}{c}{}$ can be tuned for different applications \cite{GP_learning}, this approach can be extended for nonlinear problems (see Sec.\,\ref{sec: gp_wnoj}) and enables more accurate state interpolation \cite{ fgo_zhang, Dong20174DCM}.  

\subsection{GP-WNOJ Motion Prior Model} \label{sec: gp_wnoj}
Following the approach in \cite{WNOJ}, a GP motion prior for $SE(3)$ can be defined as 
\begin{align}
\label{eq: gpprior}
    \begin{split}
         \myFrameVecDot{T}{}{}(t) &= \myFrameVec{\varpi}{}{}(t)^\land \myFrameVec{T}{}{}(t), \\
         \myFrameVecDot{\varpi}{}{}(t) &= \myFrameVec{w}{}{}(t),
    \end{split}
\end{align}
where the vehicle pose in the global frame is denoted as $\myFrameVec{T}{}{}(t)$, which can be calculated as $\myFrameVec{T}{}{}(t) = \exp{(\myFrameVec{\xi}{}{}(t)^\land)}$ with local pose $\myFrameVec{\xi}{}{}(t) = [\myFrameVec{\rho}{}{}(t)^T ~\myFrameVec{\phi}{}{}(t)^T]^T \in \mathbb{R}^{6}$. The vectors $\myFrameVec{\rho}{}{}(t)$ and $\myFrameVec{\phi}{}{}(t)$ represent the position and orientation of a local pose (e.g., in the body frame) \cite{BarfootTRO2014}. 

A local pose can be converted to $\mathfrak{se}(3)$ by applying the operator $(\cdot)^\land$. The operator $(\cdot)^\lor$ is the inverse of $(\cdot)^\land$ \cite{BarfootBook2024}. The vector $\myFrameVec{\varpi}{}{}(t)=[\myFrameVec{\nu}{}{}(t)^T~ \myFrameVec{\omega}{}{}(t)^T]^T \in \mathbb{R}^{6}$ represents the body-centric velocity. With this motion prior, the state of the GP motion model in a global frame is given as
\begin{align}
    \myFrameVec{x}{}{}(t) = \{\myFrameVec{T}{}{}(t)~ \myFrameVec{\varpi}{}{}(t)~  \myFrameVecDot{\varpi}{}{}(t)\} \in SE(3)\times\mathbb{R}^{12}.
\end{align}

However, the GP motion prior in \eqref{eq: gpprior} cannot be implemented directly using \eqref{eq: ltvsde} due to nonlinearity of the system dynamics. To address this problem, Anderson and Barfoot \cite{fullsteam} showed that a local linear GP prior can be defined between each state-timestamp pair, $t_i$ and $t_{i+1}$, by transforming the global pose $\myFrameVec{T}{}{}(t)$ into the local tangent frame, where a local pose $\myFrameVec{\xi}{}{}(t)$ can be calculated as
\begin{align}
   \myFrameVec{\xi}{i}{}(t) = \ln(\myFrameVec{T}{}{}(t)\myFrameVec{T}{t_i}{-1})^\lor, ~~ t_i \leq t  \leq t_{i+1}, \label{eq: global_to_local_pose}
\end{align}
where we consider the pose $\myFrameVec{T}{t_i}{}$ at the timestamp $t_i$ as a fixed parameter while formulating the local pose $\myFrameVec{\xi}{i}{}(t)$ for an arbitrary pose $\myFrameVec{T}{}{}(t)$ for $t > t_i$.

Because the motion between state-timestamp pairs, which are usually associated with high-frequent measurement timestamps (e.g., lidar at $\SI{10}{\Hz}$), is generally small, this local GP prior approximately represents a linear time-invariant (LTI) SDE, which can be driven from \eqref{eq: ltvsde} by assuming the system matrices remain constant. Thus, a local state variable of GP-WNOJ prior for $SE(3)$ can be defined as 
\begin{align}
     \myFrameVec{\gamma}{}{}(t) = [\myFrameVec{\xi}{}{}(t)^T ~ \myFrameVecDot{\xi}{}{}(t)^T ~ \myFrameVecDotDot{\xi}{}{}(t)^T]^T
\end{align}
and propagated using \eqref{eq:gp_discrete} to \eqref{eq:gp_Phi}. The time derivatives of the local pose can be calculated as
\begin{align}
    \myFrameVecDot{\xi}{}{}(t) &= \jac(\myFrameVec{\xi}{i}{}(t))^{-1}\myFrameVec{\varpi}{}{}(t), \\
    \myFrameVecDotDot{\xi}{}{}(t) &= -\frac{1}{2}(\jac(\myFrameVec{\xi}{}{}(t))^{-1}\myFrameVec{\varpi}{}{}(t))^\curlywedge\myFrameVec{\varpi}{}{}(t)+\jac(\myFrameVec{\xi}{}{}(t))^{-1}\myFrameVecDot{\varpi}{}{}(t)\label{eq: dot_dot_xi},
\end{align}
where the matrix $\jac$ is the left Jacobian of $SE(3)$ \cite{BarfootBook2024}. To calculate $\frac{d(\jac^{-1})}{dt}$ in closed form for \eqref{eq: dot_dot_xi}, we approximately formulate $\jac^{-1}\approx\myFrameVec{1}{}{}-\frac{1}{2}\myFrameVec{\xi}{}{\curlywedge}$ \cite{WNOJ}. 

The operator $(\myFrameVec{\xi}{}{})^\curlywedge$ represents the adjoint of $\myFrameVec{\xi}{}{\wedge}\in\mathfrak{se}(3)$  \cite{BarfootBook2024}, which can be calculated as
\begin{align}
    \myFrameVec{\xi}{}{\curlywedge}=\left[\begin{matrix}\myFrameVec{\rho}{}{}\\\myFrameVec{\phi}{}{}\end{matrix} \right] 
                                   = \left[\begin{matrix}\myFrameVec{\phi}{}{\wedge} & \myFrameVec{\rho}{}{\wedge}\\
                                                         \myFrameVec{0}{}{}          &  \myFrameVec{\phi}{}{\wedge}\end{matrix} \right].
\end{align}
Because the left Jacobian requires several matrix calculations, it can be approximated as an identity matrix $\myFrameVec{1}{}{}$ over small intervals to improve the computation efficiency\footnotemark[5] \cite{gp-slam}. 
\footnotetext[5]{We implement this trick as a configuration in the proposed \texttt{gnssFGO}.}

Given a local state variable that represents the origin system state for each state-timestamp pair, we can retrieve the WNOJ motion model for two successive local state variables in the local frame as
\begin{align}
\label{eq: gp_gamma}
    \begin{split}
        \myFrameVec{\gamma}{i}{}(t_i) &= [\myFrameVec{0}{}{} ~\myFrameVec{\varpi}{}{}(t)^T ~\myFrameVecDot{\varpi}{}{}(t)^T]^T, \\
        \myFrameVec{\gamma}{i}{}(t_{i+1}) &= \left[\begin{matrix}
                                                    \ln(\myFrameVec{T}{i+i,i}{})^\lor\\
                                                    \jac_{i+1}^{-1}\myFrameVec{\varpi}{i+1}{} \\
                                                    -\frac{1}{2}(\jac_{i+1}^{-1}\myFrameVec{\varpi}{i+1}{})^\curlywedge\myFrameVec{\varpi}{i+1}{} + \jac_{i+1}^{-1}\myFrameVecDot{\varpi}{i+1}{}
                                                    \end{matrix} \right].
    \end{split}
\end{align}

Using the GP-WNOJ prior, a state at an arbitrary time $\tau \in (i, i+1)$ can be queried as
\begin{align}
\label{eq: gp_query}
    \begin{split}
        \myFrameVec{T}{\tau}{} &= \exp\Big \{ [ \myFrameVec{\Lambda}{12}{}(\tau)\myFrameVec{\varpi}{i}{} + \myFrameVec{\Omega}{13}{}(\tau)\myFrameVecDot{\varpi}{i}{} + \myFrameVec{\Sigma}{11}{}(\tau)\ln(\myFrameVec{T}{i+1,i}{})^\lor +\\
                               & + \myFrameVec{\Omega}{12}{}(\tau)\jac_{i+1}^{-1}\myFrameVec{\varpi}{i+1}{} + \\
                               & + \myFrameVec{\Omega}{13}{}(\tau)( -\frac{1}{2}(\jac_{i+1}^{-1}\myFrameVec{\varpi}{i+1}{})^\curlywedge\myFrameVec{\varpi}{i+1}{} + \jac_{i+1}^{-1}\myFrameVecDot{\varpi}{i+1}{})]^\land \Big\} \myFrameVec{T}{i}{},
    \end{split}
\end{align}
where $\myFrameVec{\Lambda}{}{}$ and $\myFrameVec{\Omega}{}{}$ are vehicle transition matrices obtained from \eqref{eq:gp_discrete} to \eqref{eq:gp_Q}. 

\begin{remark} 
    \textit{\textbf{Hyper-Parameter of GP-WNOJ model:}} As discussed in \cite{WNOJ}, representing a realistic system transition using the GP motion priors requires proper tuning of the power spectral density matrix $\myFrameVec{Q}{c}{}$. In this work, we assume that $\myFrameVec{Q}{c}{}$ is a constant diagonal matrix defined as $\myFrameVec{Q}{c}{} = \mathrm{diag}(\myFrameVec{q}{c}{})$ with a 6-D hyper-parameter $\myFrameVec{q}{c}{}$. The $\myFrameVec{q}{c}{}$ was manually tuned for the experiments. For more details, see Sec.\,\ref{sec: res_wnoj}.
\end{remark}

\subsection{Measurement Models} \label{sec: meas_models}

\subsubsection{GNSS Observations} \label{sec: meas_gnss}
Generally, a single antenna GNSS receiver can provide both position, velocity, and time (PVT) solutions and raw observations. A GNSS receiver equipped with multiple antennas or an inertial sensor can also produce a position, velocity, and altitude (PVA) solution. In this work, we only use the GNSS-PVT solution from a low-cost in the loosely coupled fusion because it does not require additional hardware components. The altitude in the GNSS-PVA solution is taken as a reference. As the pose and velocity of the GNSS solution can be directly associated with the state variables in the FGO, we only present the measurement models for the raw GNSS observations: pseudorange $\rho$ and Doppler-shift $\Delta\myFrameScalar{f}{k}{}$ in this section.

In localization approaches that tightly fuse the GNSS observations, pseudorange and Doppler-shift are commonly used and well studied \cite{groves}. The pseudorange $\myFrameScalar{\rho}{}{}$ represents a geometric distance between the phase center of the GNSS antenna and the associated satellite, which contains several range delays due to satellite orbit bias and atmospheric delays. The pseudorange can be modeled with respect to the antenna position as
\begin{equation}
    \myFrameScalar{\rho}{k}{}= \underbrace{\left\|\myFrameVec{p}{\rm ant}{e}-\myFrameVec{p}{\mathrm{sat},k}{e}\right\|}_{\text{1-D geometric range}}
                        + \myFrameScalar{c}{b}{} - c_{b, \mathrm{sat}} + T + I + M + w_{\rho,k}\mathrm{,}
\label{eq: rho}
\end{equation}
where the vectors $\myFrameVec{p}{a}{e}$ and $\myFrameVec{p}{\mathrm{sat},k}{e}$ represent the positions of the GNSS antenna and $k$-th satellite in ECEF frame\footnotemark[6], respectively. The variables $\myFrameScalar{c}{b}{}$ and $c_{b, \mathrm{sat}}$ represent the bias due to receiver clock delay and satellite clock delay. The tropospheric, ionospheric, and multipath delays are denoted as $T(t)$, $I(t)$ and $M$, respectively. The pseudorange noise is $w_{\rho}$.

\footnotetext[6]{We represent the vehicle and satellite position in the ECEF frame instead of the Earth-centered inertial frame for clarity by assuming that the GNSS pre-processing has calibrated the earth rotation during GNSS signal propagation.}
The Doppler-shift\footnotemark[7] $\Delta\myFrameScalar{f}{k}{}$ measures the difference in frequency between the original and received carrier signal of a satellite, which is usually obtained in the carrier-phase tracking loop \cite{groves}. With this observation, the vehicle velocity related to the satellite velocity can be represented as
\begin{equation}
    -\lambda_c\Delta\myFrameScalar{f}{k}{} =  \underbrace{(\myFrameVec{u}{\mathrm{ant}}{\rm sat})^T (\myFrameVec{v}{\rm ant}{e}-\myFrameVec{v}{\mathrm{sat},k}{e})}_{\text{1-D range rate}} + \myFrameScalar{c}{d}{} - \myFrameScalar{c}{d, \mathrm{sat}}{} + w_{\Delta\myFrameScalar{f}{k}{}}.
\label{eq: delta_rho}
\end{equation}
In \eqref{eq: delta_rho}, the constant $\lambda_c$ is the wavelength of the GNSS signal. The unit vector $\myFrameVec{u}{\mathrm{ant}}{\rm sat}$ represents the direction from the antenna to the $k$-th satellite. We denote the satellite and antenna velocities in the ECEF frame\footnotemark[6] by $\myFrameVec{v}{\mathrm{sat},k}{e}$ and $\myFrameVec{v}{\rm ant}{e}$, respectively. The receiver clock drift and the satellite clock drift are given as $\myFrameScalar{c}{d}{}$ and $\myFrameScalar{c}{d,\mathrm{sat}}{}$. We use the scalar $w_{\Delta\myFrameScalar{f}{k}{}}$ to denote the noise of the measured Doppler-shift.

To formulate the pseudorange and Doppler-shift factors, we assume that the satellite clock delay $c_{b, \mathrm{sat}}$ and drift $\myFrameScalar{c}{d, \mathrm{sat}}{}$ are eliminated using the received navigation messages in a GNSS pre-processing process \cite{fgo_zhang}. We used well-calibrated correction data from a reference station to cancel the tropospheric and ionospheric interference. The multipath error $M$ is not explicitly modeled in this work as m-estimators are used to reject measurement outliers, see Sec.\,\ref{sec: noise_model}. We filter out all GNSS observations from satellites with an elevation angle less than \SI{15}{\degree}\footnotemark[8].

\footnotetext[7]{In some literature, the Doppler-shift is also formulated as deltarange or pseudorange rate measurement\cite{groves, gnss_handbook}. A positive Doppler-shift denotes that the receiver is approaching the tracked satellite.}

\footnotetext[8]{This is an ad-hoc choice, generally used in ground vehicle navigation approaches and GNSS receivers.}

\subsubsection{IMU Pre-Integration}
In graph-optimization-based state estimation approaches, the IMU pre-integration, introduced in \cite{imu_pre_se3}, is generally utilized to integrate high-frequency IMU measurements as between-state factors for the optimization procedures running at a lower rate. The pre-integrated IMU measurements represent the relative motion increments on manifold. These relative motion increments can be assumed unchanged while re-linearizing the consecutive state variables in the optimization iterations, resulting in efficient computation. Due to limited space, we refer the reader to \cite{imu_pre_se3} for more details. We use this IMU mechanism to formulate the IMU factor, as presented in Sec.\,\ref{sec: factors_imu}.

\subsubsection{Lidar Odometry} \label{method: lidarodom}
We adapt the feature extraction and matching methods from a feature-based lidar odometry \cite{loam, liosam} to obtain the relative motion increments between two laser keyframes. The coordinates of raw lidar points acquired in different timestamps are re-calibrated using the IMU measurement to the original timestamp of the lidar scan. We classify the calibrated points into edge and planar features, $\myFrameVec{F}{t}{}=\{\myFrameVec{F}{t}{e}~\myFrameVec{F}{t}{p}\}$, based on the smoothness metric shown in \cite{loam, legoloam}. In scan registration, all $k$ features in $\myFrameVec{F}{t+1}{}$ of the current scan are associated with pose priors $\myFrameVec{T}{t+1,1:k}{w}$ and used to find the best transformation $\Delta\myFrameVec{T}{t, t+1}{w}$ from the last laser scan by solving an optimization problem that takes the distance between the corresponding features in $\myFrameVec{F}{t}{}$ using a Gauss-Newton algorithm.

In \cite{liosam}, a lidar-centric SLAM approach is presented that optionally fuses the GNSS positioning solution. This approach can only present accurate state estimates if the scan registration converges and sufficient global references (e.g., GNSS position or loop closure) are available. In contrast to \cite{liosam}, we query the vehicle states at scan timestamps from a previously built time-centric graph and integrate the transformation $\Delta\myFrameVecTilde{T}{t, t+1}{w}$ as between-pose constraints, which is used to formulate the between-pose factor (see Sec.\,\ref{sec: factor_bp}). After the graph optimization, we query the optimized states again using the GP motion model and update lidar keyframe poses in frame $\myFrameScalar{w}{}{}$ using the following transformation
\begin{align}
\label{eq: lidar_p}
    \myFrameVec{T}{l,t}{w} = \myFrameVec{T}{l,\rm anc}{e,-1}\myFrameVec{T}{l,t}{e},                         
\end{align} 
where the transformation matrices $\myFrameVec{T}{l,t}{e}$ and $\myFrameVec{T}{l,t}{w}$ denote lidar poses in frame $\myFrameScalar{e}{}{}$ and frame $\myFrameScalar{w}{}{}$, respectively. As lidar odometry requires a state-space representation in a local-world (aka local-tangent) frame $\myFrameScalar{w}{}{}$ where the $z$-axis is gravity aligned, we query an anchor pose $\myFrameVec{T}{l,\rm anc}{e} = \Big[\begin{smallmatrix} \myFrameVec{R}{l,\rm anc}{e}&\myFrameVec{p}{l,\rm anc}{e}\\ \myFrameVec{0}{}{}&1 \end{smallmatrix}\Big]$ of the lidar sensor on first scan and initialize a local-world frame of the lidar odometry by setting the anchor pose as its origin. In contrast to \cite{gvins}, a coarse orientation estimate is unnecessary in our work to align the local-world frame and the navigation frame because a prior vehicle heading is provided by the dual-antenna GNSS receiver.

\subsubsection{Optical Speed Sensor}
We employ a high-grade vehicle optical speed sensor that provides unbiased 2-D velocity observations $\myFrameVecBar{v}{t}{b}$ in the body frame at $\SI{100}{\Hz}$. The 2-D velocity observations can be associated with the vehicle velocity in the state vector using
\begin{align}
    \label{eq: vel_logger}
    \myFrameVecTilde{v}{t}{b} = \left[\begin{matrix}
                                      \myFrameScalarTilde{v}{t,x}{b}\\
                                      \myFrameScalarTilde{v}{t,y}{b}
                                      \end{matrix} \right]
                              = \left[\begin{matrix}
                                  1 & 0 & 0\\
                                  0 & 1 & 0\\
                              \end{matrix}\right] \cdot \myFrameVec{R}{e}{b}\myFrameVec{v}{b}{e},
\end{align}
where the vector $\myFrameVecTilde{v}{t}{b}$ represents the observed 2-D velocity components in frame $\myFrameScalar{b}{}{}$, which can be evaluated with the vehicle velocity variable $\myFrameVec{v}{b}{e}$ transformed with the inverse rotation matrix $\myFrameVec{R}{e}{b}$ back to frame $\myFrameScalar{b}{}{}$.

\section{FGO for Vehicle Localization} \label{sec: loca}
This section presents our implementation of the proposed \texttt{gnssFGO} for two sensor fusion schemes. In loosely coupled fusion, we fuse the PVT solution from a low-cost GNSS receiver with the IMU measurements, the observed 2-D vehicle velocity from a high-grade speed sensor, and the lidar odometry. To defend the superiority of fusing raw GNSS observations for vehicle localization, we propose a tightly coupled fusion of raw GNSS observations with IMU measurements and lidar odometry, which is evaluated with the baseline trajectory. In this section, we introduce all probabilistic factor formulations and the proposed factor graph structures.

\subsection{State Variables}  \label{sec: state}
The state variable at timestamp $t$ in this work is defined as
\begin{align}
    \myFrameVec{x}{t}{} \triangleq \{\myFrameVec{T}{b,t}{e} ~ \myFrameVec{v}{b,t}{e} ~ \myFrameVec{b}{b,t}{\mathrm{a}} ~ \myFrameVec{b}{b,t}{\mathrm{g}} ~ \myFrameVec{c}{t}{\mathrm{r}}\}.
\end{align}
We estimate the vehicle pose $\myFrameVec{T}{b,t}{e} \in SE(3)$ and 6-D velocity $\myFrameVec{v}{b}{e}$ in frame $\myFrameScalar{e}{}{}$. The vectors $\myFrameVec{b}{b,t}{\rm a}$ and $\myFrameVec{b}{b,t}{\rm g}$ denote the 3-D biases of the accelerometer and gyroscope, respectively. The 2-D vector $\myFrameVec{c}{t}{\rm r} = [\myFrameScalar{c}{b,t}{}~\myFrameScalar{c}{d,t}{}]^T$ represents the GNSS receiver clock bias $\myFrameScalar{c}{b,t}{}$ and drift $\myFrameScalar{c}{d,t}{}$, which is only estimated by the tightly coupled fusion of raw GNSS observations. 

\begin{remark}
    \textit{\textbf{Acceleration of GP-WNOJ:}} Unlike \cite{WNOJ}, we do not estimate 6-D accelerations in GP motion models to reduce the dimension of the state vector. Instead, we consider the vehicle accelerations measured by the IMU as inputs to the WNOJ model.
\end{remark}

\subsection{Factor Formulations}
\subsubsection{Pre-Integrated IMU Factor} \label{sec: factors_imu}
 Following \cite{imu_pre_se3}, we define the error function of the IMU factor between two consecutive state variables at timestamps $t_i,~t_j$ as
\begin{align}
    \left\|\myFrameVec{e}{ij}{\mathrm{imu}}\right\|^2= 
    \left\|[\myFrameVec{r}{\Delta\myFrameVec{R}{ij}{}}{T}
    ~\myFrameVec{r}{\Delta\myFrameVec{v}{ij}{}}{T}
    ~\myFrameVec{r}{\Delta\myFrameVec{p}{ij}{}}{T}]^T \right\|_{\myFrameVec{\Sigma}{}{\mathrm{imu}}}^2,
\end{align}
where
\begin{align}
    \myFrameVec{r}{\Delta\myFrameVec{R}{ij}{}}{}&=\mathrm{Log}(\Delta\myFrameVecTilde{R}{ij}{}(\myFrameVec{b}{i}{\rm g}))\myFrameVec{R}{i}{T}\myFrameVec{R}{i}{}, \label{eq: imu_r} \\
    \myFrameVec{r}{\Delta\myFrameVec{v}{ij}{}}{}&=\myFrameVec{R}{i}{T}(\myFrameVec{v}{j}{}-\myFrameVec{v}{i}{}-\myFrameVec{g}{}{}\Delta t_{ij})-\Delta\myFrameVecTilde{v}{ij}{}(\myFrameVec{b}{i}{\rm g},\myFrameVec{b}{i}{\rm a}), \\
    \myFrameVec{r}{\Delta\myFrameVec{p}{ij}{}}{} &=\myFrameVec{R}{i}{T}(\myFrameVec{p}{j}{}-\myFrameVec{p}{i}{}-\myFrameVec{v}{i}{}\Delta \myFrameScalar{t}{ij}{}-\frac{1}{2}\myFrameVec{g}{}{}\Delta\myFrameScalar{t}{ij}{2})-\Delta\myFrameVecTilde{p}{ij}{}(\myFrameVec{b}{i}{\rm g},\myFrameVec{b}{i}{\rm a}). \label{eq: imu_p}
\end{align}
In \eqref{eq: imu_r} to \eqref{eq: imu_p}, we omit the bias derivatives that can be ignored between two state variables. The motion increments $\{\Delta\myFrameVecTilde{R}{ij}{}~\Delta\myFrameVecTilde{v}{ij}{}~\Delta\myFrameVecTilde{p}{ij}{}\}$ are provided by the IMU pre-integration with
\begin{align}
    \label{eq:delta_R}
    \Delta\myFrameVecTilde{R}{ij}{} &= \prod_{k=i}^{j-1}\mathrm{Exp}((\myFrameVecTilde{\omega}{k}{} - \myFrameVec{b}{i}{\rm g} - \myFrameVec{\eta}{k}{\rm g})\Delta t),\\
    \label{eq:delta_v}
    \Delta\myFrameVecTilde{v}{ij}{} &= \sum_{k=i}^{j-1} \Delta\myFrameVecTilde{R}{ik}{}(\myFrameVecTilde{a}{k}{}-\myFrameVec{b}{i}{\rm a}- \myFrameVec{\eta}{k}{\rm a})\Delta t,\\
    \label{eq:delta_p}
    \Delta\myFrameVecTilde{p}{ij}{} &= \sum_{k=i}^{j-1} \Big[\Delta\myFrameVec{v}{ik}{}\Delta t + \frac{1}{2}\Delta\myFrameVecTilde{R}{ik}{}(\myFrameVecTilde{a}{k}{}-\myFrameVec{b}{i}{\rm a}- \myFrameVec{\eta}{k}{\rm a})\Delta t^2\Big],
\end{align}
where the raw vehicle acceleration $\myFrameVecTilde{a}{}{}$ and rotation rate $\myFrameVecTilde{\omega}{}{}$ from the IMU are integrated.
The pre-defined noise parameters $\{\myFrameVec{\eta}{\rm a}{}~\myFrameVec{\eta}{\rm g}{}\}$ are propagated to acquire the covariance matrix $\myFrameVec{\Sigma}{}{\mathrm{imu}}$ \cite{imu_pre_se3}. The gravity vector is updated according to the current position in the $e$ frame for each pre-integration.

As in \cite{imu_pre_se3}, we estimate the accelerometer and gyroscope biases with the Brownian motion model by formulating the bias error function as
\begin{align}
    \left\|\myFrameVec{e}{ij}{b}\right\|^2= 
    \left\|\myFrameVec{b}{j}{\rm a} - \myFrameVec{b}{i}{\rm a}\right\|_{\myFrameVec{\Sigma}{}{\rm ba}}^2 + 
    \left\|\myFrameVec{b}{j}{\rm g} - \myFrameVec{b}{i}{\rm g}\right\|_{\myFrameVec{\Sigma}{}{\rm bg}}^2.
\end{align}

\subsubsection{Between-Pose Factor}\label{sec: factor_bp}
For the relative odometry observations $\Delta\myFrameVecTilde{T}{i, j}{e} = \{\Delta\myFrameVecTilde{R}{i, j}{e}~\Delta\myFrameVecTilde{p}{i, j}{e}\}$, we follow the original implementation in \cite{theory_betweenpose} and formulate the between pose factor represented as
\begin{equation}
\label{eq: err_between}
    \left\|\myFrameVec{e}{i, j}{\mathrm{bp}}\right\|^2 = 
    \left\| 
            \ln(\myFrameVec{T}{i}{e,-1}\myFrameVec{T}{j}{e}\Delta\myFrameVecTilde{T}{i, j}{e})^\vee
    \right\|_{\myFrameVec{\Sigma}{}{\mathrm{bp}}}^2,
\end{equation}
where the pose $\myFrameVec{T}{i}{e}$ and $\myFrameVec{T}{j}{e}$ are queried using timestamps associated with two successive lidar scans.

\subsubsection{Velocity Factor}
We use the 2-D observations $\myFrameVecTilde{v}{t}{b}$ to formulate the navigation velocity factor. As the measured velocity can be directly associated with the velocity in state variables, as denoted in \eqref{eq: vel_logger}, we formulate the error function for the velocity observations considering the lever arm $\myFrameVec{l}{}{b,\mathrm{vel}}$ from the body frame to the sensor center as
\begin{align}
    \left\|\myFrameVec{e}{i}{\mathrm{vel}}\right\|^2 =  
    \left\|\left[\begin{matrix}
                                  1 & 0 & 0\\
                                  0 & 1 & 0\\
                              \end{matrix}\right] \cdot (\myFrameVec{R}{e,i}{b}\myFrameVec{v}{b,i}{e} +\myFrameVec{\omega}{i}{b\wedge} \myFrameVec{l}{}{b,\mathrm{vel}}) - \myFrameVecTilde{v}{i}{b}
    \right\|_{\myFrameVec{\Sigma}{}{\mathrm{vel}}}^2.
\end{align}

\subsubsection{GNSS-PVT Factor}\label{sec: factor_pvt}
We propose a generalized implementation of the GNSS-PVT factor for the observed antenna position $\myFrameVecTilde{p}{\mathrm{ant}}{e}$ and the velocity $\myFrameVecTilde{v}{\mathrm{ant}}{n}$. Taking into account the lever arm $\myFrameVec{l}{\mathrm{ant}}{b}$ from the IMU center to the phase center of the GNSS antenna, we calculate the antenna position at timestamp $t_i$ as $\myFrameVec{p}{\mathrm{ant},i}{e}=\myFrameVec{p}{b,i}{e} + \myFrameVec{R}{b,i}{e}\myFrameVec{l}{\mathrm{ant}}{b}$ and velocity as $\myFrameVec{v}{\mathrm{ant},i}{e}=\myFrameVec{v}{b,i}{e} + \myFrameVec{R}{b,i}{e}(\myFrameVec{\omega}{i}{b})^\wedge\myFrameVec{l}{\mathrm{ant}}{b}$. Thus, the error function can be derived as
\begin{align}
    \left\|\myFrameVec{e}{i}{\mathrm{pvt}}\right\|^2= 
    \left\|[\myFrameVec{r}{\myFrameVec{p}{i}{}}{T}
    ~\myFrameVec{r}{\myFrameVec{v}{i}{}}{T}]^T \right\|_{\myFrameVec{\Sigma}{}{\mathrm{pvt}}}^2,
\end{align}
with
\begin{align}
    \myFrameVec{r}{\myFrameVec{p}{i}{}}{} &= \myFrameVec{p}{\mathrm{ant},i}{e} - \myFrameVecTilde{p}{\mathrm{ant},i}{e}, \\
    \myFrameVec{r}{\myFrameVec{v}{i}{}}{} &= \myFrameVec{R}{e,i}{n}\myFrameVec{v}{\mathrm{ant},i}{e} - \myFrameVecTilde{v}{\mathrm{ant},i}{n},
\end{align}
where the rotation matrix $\myFrameVec{R}{e,i}{n}$ is given in \eqref{eq: R_e_n} by substituting the geodetic coordinate of the main antenna $\myFrameVec{p}{\mathrm{ant,i}}{\rm LLH}$ as the origin. We use the measured standard deviations in the GNSS solutions to formulate the covariance matrix $\myFrameVec{\Sigma}{}{\mathrm{pvt}}$.

\subsubsection{Pseudorange and Doppler-shift (PrDo) Factor}\label{sec: factor_PrDo}
We derive the error function for the pre-processed pseudorange and Doppler-shift observations with \eqref{eq: rho} and \eqref{eq: delta_rho} as
\begin{equation}
     \left\|\myFrameVec{e}{i}{\mathrm{PrDo}}\right\|^2 
     =\left\|[\myFrameScalar{r}{i}{\mathrm{Pr}} ~ \myFrameScalar{r}{i}{\mathrm{Do}}]^T\right\|_{\myFrameVec{\Sigma}{}{\mathrm{PrDo}}}^2,
    \label{eq: e_gnss}
\end{equation}
where 
\begin{align}
    \myFrameScalar{r}{i}{\mathrm{Pr}} &= \left\|\myFrameVec{p}{\mathrm{ant},i}{e} - \myFrameVec{p}{\mathrm{sat},k,i}{e}\right\| + c_{b,i}  - \myFrameScalar{\rho}{k,i}{}, \label{eq: e_rho}\\
    \myFrameScalar{r}{i}{\mathrm{Do}} &= (\myFrameVec{u}{\mathrm{ant},i}{\rm sat})^T\left(\myFrameVec{v}{\mathrm{ant},i}{e}-\myFrameVec{v}{\mathrm{sat},k,i}{e}\right)+c_{d,i} + \lambda_c\Delta\myFrameScalar{f}{k,i}{}. \label{eq: e_nu}
\end{align}

We consider a scaled carrier-to-noise ratio $(C/N_0)$ with hyper-parameters $\lambda_{\rho}$ and $\lambda_{\Delta f_k}$ to represent the variance of pseudorange and Doppler-shift observations, which is denoted as
\begin{align}
    \label{eq: cn0}
    \myFrameScalar{\eta}{\rho}{2} = \lambda_{\rho} 10^{-\frac{\mathrm{C/N}_0}{10}} ~ \text{and} ~ \myFrameScalar{\eta}{\Delta f}{2} = \lambda_{\Delta f} 10^{-\frac{\mathrm{C/N}_0}{10}}. ~
\end{align}

\subsubsection{GNSS Receiver Clock Error Factor}
In the tight coupling of the raw GNSS observations, the unknown receiver clock bias and drift (cbd) are estimated in the state variable by assuming a constant drifting model, which can be fused as
\begin{equation}
\label{eq:e_clock}
    \left\|\myFrameVec{e}{i}{\mathrm{cbd}}\right\|^2 = \left\| 
    \left[ \begin{matrix} 
    1  & \Delta t \\
    0  &  1
    \end{matrix}\right]\Big[\begin{matrix}
        c_{b,i-1} \\ c_{d,i-1}
    \end{matrix}\Big] - \Big[\begin{matrix}
        c_{b,i} \\ c_{d,i}
    \end{matrix}\Big] \right\|_{\myFrameVec{\Sigma}{}{\mathrm{cbd}}}^2.
\end{equation}

\subsubsection{GP-WNOJ Motion Prior Factor}\label{sec: factor_gp_motion}
We implement the GP-WNOJ motion model as between-state factors, similar to \cite{gp-dong}. The error function was originally given in \cite{WNOJ} using \eqref{eq: gp_gamma}. We summarize this error function for convenience as
\begin{align}
    \label{eq: gp_prior_err}
    \left\|\myFrameVec{e}{ij}{\mathrm{gp}}\right\|^2= 
    \left\|[\myFrameVec{r}{\Delta\myFrameVec{\gamma}{ij}{}}{T}
    ~\myFrameVec{r}{\Delta\myFrameVec{\varpi}{ij}{}}{T}]^T \right\|_{\myFrameVec{\Sigma}{}{\mathrm{gp}}}^2,
\end{align}
where
\begin{align}
  \label{eq: gp_prior_err_vel}
    \myFrameVec{r}{\Delta\myFrameVec{\gamma}{ij}{}}{} &= \ln(\myFrameVec{T}{j,i}{})^\lor - (t_{j}-t_i)\myFrameVec{\varpi}{i}{} - \frac{1}{2}(t_{j}-t_i)^2\myFrameVecDot{\varpi}{i}{}, \\
    \myFrameVec{r}{\Delta\myFrameVec{\varpi}{ij}{}}{} &= \jac_{j,i}^{-1}\myFrameVec{\varpi}{j}{} - \myFrameVec{\varpi}{i}{} - (t_j-t_i)\myFrameVecDot{\varpi}{i}{}.
\end{align}
As introduced in Sec.\,\ref{sec: state}, we used the measured accelerations of the IMU in our GP motion models. Thus, only the 6-D pose and the 6-D velocity are evaluated in GP-WNOJ motion factors, so that $\myFrameVec{e}{ij}{\mathrm{gp}} \in \mathbb{R}^{12}$. The analytical Jacobians of the GP motion models can be found in \cite{phd_sean, WNOJ}.

\subsection{Loosely Coupled FGO} \label{sec: lc}
Although the loosely coupled fusion with GNSS and IMU measurements has been shown to be less performant compared to tight coupling \cite{ekf_fgo_compare}, we implemented a loosely coupled fusion of sensor observations, including the 2-D speed sensor and the lidar odometry, to i) study the performance gain by fusing multiple sensor observations; ii) evaluate the loosely and tightly coupled fusion for GNSS-based vehicle localization in challenging areas; iii) demonstrate the flexibility and scalability of the proposed method.

The proposed factor graph is shown in Fig.\,\ref{fig: fg_lc}. The states $\myFrameVec{x}{1:t}{}$ are created deterministically on the graph independently of any measurement. If a measurement cannot be associated with any state variable, a state $\myFrameVecHat{x}{i+\tau}{}$ between two state variables $\myFrameVecHat{x}{i}{}$ and $\myFrameVecHat{x}{i+1}{}$ (where $t_i < \tau < t_{i+1}$) is queried for the error evaluation.

The optimization problem can then be formulated as
\begin{equation}
    \begin{split}
       \myFrameVecHat{x}{}{} &= \argminD_{\myFrameVec{x}{}{}}\Big(
         \left\|\myFrameVec{e}{}{0}\right\|_{\myFrameVec{\Sigma}{0}{}}^{2}  +
         \sum_{i=1}^{M}\left\|\myFrameVec{e}{i}{\mathrm{imu}}\right\|_{\myFrameVec{\Sigma}{}{\mathrm{imu}}}^{2} +
         \sum_{i=1}^{M}\left\|\myFrameVec{e}{i}{\mathrm{gp}}\right\|_{\myFrameVec{\Sigma}{}{\mathrm{gp}}}^{2} +  \\ 
       &+\sum_{i=1}^{N}\left\|\myFrameVec{e}{i}{\mathrm{vel}}\right\|_{\myFrameVec{\Sigma}{}{\mathrm{vel}}}^{2} 
        +\sum_{i=1}^{K}\left\|\myFrameVec{e}{i}{\mathrm{pvt}}\right\|_{\myFrameVec{\Sigma}{}{\mathrm{pvt}}}^{2}
        +\sum_{i=1}^{J}\left\|\myFrameVec{e}{i}{\mathrm{bp}}\right\|_{\myFrameVec{\Sigma}{}{\mathrm{bp}}}^{2}\Big),
    \end{split}
    \label{eq: fg_lc_map}
\end{equation}
where the error term $\myFrameVec{e}{}{0}$ represents the prior factor obtained at initialization or from marginalization. Because sensor observations are received asynchronously other than estimation timestamps $M$, we use different index notations $N,~ K,~J$ to indicate the number of sensor observations in \eqref{eq: fg_lc_map}. 

\subsection{Tightly Coupled FGO} \label{sec: tc}
\begin{figure}[!t]
    \centering
    \includegraphics[width=0.45\textwidth]{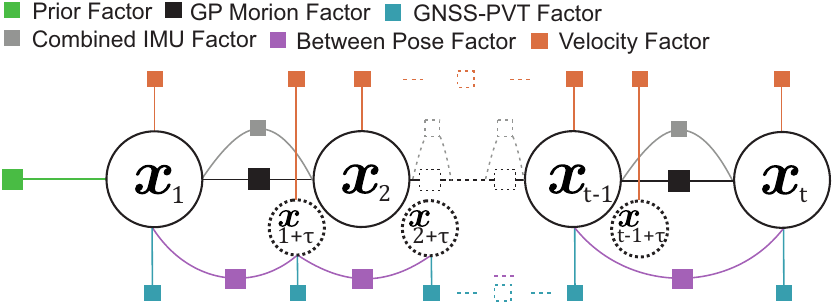}
    \caption{A general graph of loose coupling in \texttt{gnssFGO}.}
    \label{fig: fg_lc}
\end{figure}

\begin{figure}[!t]
    \centering
    \includegraphics[width=0.43\textwidth]{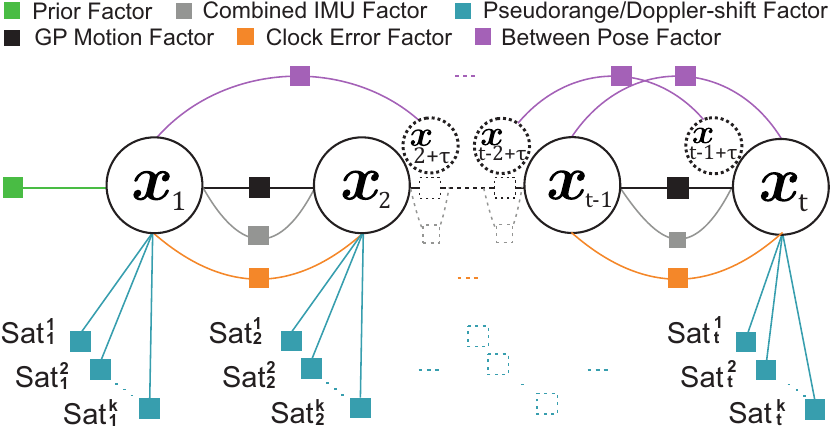}
    \caption{A general graph of tight coupling in \texttt{gnssFGO} with $k$ raw satellite observations at each timestamp $t$ denoted as $\text{Sat}_t^k$.}
    \label{fig:fg_proposed}
\end{figure}

In contrast to the loosely coupled fusion approach, a tightly coupled fusion of raw GNSS observations contributes more constraints with multiple observed satellites to state variables, as illustrated in Fig.\,\ref{fig:fg_proposed}. Unlike Fig.\,\ref{fig: fg_lc}, we include the pseudorange and Doppler-shift factors in the graph, providing redundant constraints to each state variable. To improve the robustness while GNSS observations are degraded or lost in challenging areas, we include lidar odometry as between-state constraints to improve the consistency of the estimated trajectory. The receiver clock error factor is also added to the graph. In this fusion mechanism, we do not fuse the measurements from the 2-D velocity, which is not commonly used in vehicle localization approaches, aiming to highlight the robustness of the tightly coupled fusion (see discussion in Sec.\,\ref{sec: res_lc}) and Sec.\,\ref{sec: res_tc}). 

The optimization problem with sensor observations from different time domains becomes
\begin{equation}
    \begin{split}
       \myFrameVecHat{X}{}{} =\argminD_{\myFrameVec{x}{}{}}\Big(
        &\left\|\myFrameVec{e}{i}{0}\right\|_{\myFrameVec{\Sigma}{0}{}}^{2} + \sum_{i=1}^{M}\left\|\myFrameVec{e}{i}{\mathrm{IMU}}\right\|_{\myFrameVec{\Sigma}{}{\mathrm{IMU}}}^{2} +
        \sum_{i=1}^{M}\left\|\myFrameVec{e}{i}{\mathrm{gp}}\right\|_{\myFrameVec{\Sigma}{}{\mathrm{gp}}}^{2} +\\
        &+ \sum_{i=1}^{N}\left\|\myFrameVec{e}{i}{\mathrm{bp}}\right\|_{\myFrameVec{\Sigma}{}{\mathrm{bp}}}^{2} + \sum_{i=1}^{M}\left\|\myFrameVec{e}{i}{\mathrm{cbd}}\right\|_{\myFrameVec{\Sigma}{}{\mathrm{cbd}}}^{2} +\\
        &+ \sum_{i=1}^{J}\sum_{s=1}^{K}\left\|\myFrameVec{e}{s,i}{\mathrm{PrDo}}\right\|_{\myFrameVec{\Sigma}{}{\mathrm{PrDo}}}^{2}\Big).
    \end{split}
    \label{eq:fg_map}
\end{equation}

\subsection{Noise Models}\label{sec: noise_model}
In this work, we formulate the covariance matrices of the GNSS-related factors using noise values provided in GNSS observations, as presented in Sec.\,\ref{sec: factor_pvt} and Sec.\,\ref{sec: factor_PrDo}. The IMU noise is characterized by the Allan noise parameters\footnotemark[9], which is used to calculate $\myFrameVec{\Sigma}{}{\mathrm{imu}} \in \mathbb{R}^{6\times6}$, $\myFrameVec{\Sigma}{}{\mathrm{ba}}\in \mathbb{R}^{3\times3}$, and $\myFrameVec{\Sigma}{}{\mathrm{bg}}\in \mathbb{R}^{3\times3}$. Because there are no noise indicators for lidar odometry, speed sensor, and receiver clock errors, we formulate the covariance matrices $\myFrameVec{\Sigma}{}{\mathrm{bp}}\in \mathbb{R}^{6\times6}$, $\myFrameVec{\Sigma}{}{\mathrm{vel}}\in \mathbb{R}^{2\times2}$, and $\myFrameVec{\Sigma}{}{\mathrm{cbd}}\in \mathbb{R}^{2\times2}$ manually as diagonal matrices using ad hoc parameters. Noise models of different factors can be configured to use m-estimators \cite{BarfootBook2024}. In our experiments, we use the m-estimator with \textit{Cauchy} loss \cite{robust_statistics} in the factors, such as GNSS-PVT factors, which may be affected by outlier measurement due to strong corruption in urban areas.

\footnotetext[9]{\url{https://github.com/ori-drs/allan_variance_ros}}
\subsection{System Overview} \label{sec: sys_overview}
\begin{figure}[!t]
    \centering
    \includegraphics[width=0.49\textwidth]{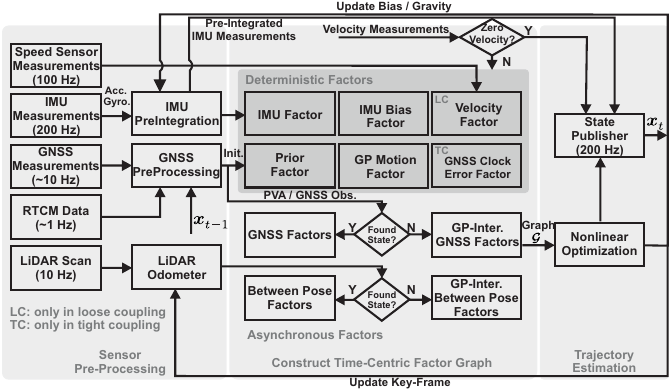}
    \caption{System overview showing all data interfaces and factor types by implementing Alg.\,\ref{alg: TC-FGO}.}
    \label{fig: sys}
\end{figure}

The system overview with the implementation of Alg.\,\ref{alg: TC-FGO} and all data interfaces is shown in Fig.\,\ref{fig: sys}. The sensor data are received and pre-processed in separate processes. We construct the time-centric factor graph in a two-stage process, as introduced in Alg.\,\ref{alg: TC-FGO}. The first stage (line 4-10 of Alg.\,\ref{alg: TC-FGO}) includes between-state factors and delay-free IMU factors to build a deterministic graph on time. Subsequently, asynchronous sensor observations are fused into the deterministic graph by aligning the timestamps between the measurement and the state variables (line 11-24 of Alg.\,\ref{alg: TC-FGO}). For measurements that cannot be aligned with any state, two successive state variables are queried to construct a GP-interpolated state for measurement evaluation in optimization procedures. The time-centric graph can be optimized using a fixed-lag batch optimizer \cite{fixedlagRanganathan} or a fixed-lag incremental smoother iSAM2 \cite{kaess_isam2} at a lower frequency. In the experimental results, the estimated trajectories in the error metrics are optimized using iSAM2. We also evaluate both smoothers with respect to both estimator performance and computation efficiency, as presented in Sec.\,\ref{sec: res_smoothers}. After each optimization procedure, we forward the optimized state variables to a state publisher and sensor pre-processing modules. The state publisher is associated with the IMU sensor and provides high-frequent state estimates at $\SI{200}{Hz}$.

\begin{remark} \label{rem: near_zero_velocity}
     \textbf{Near-Zero-Velocity Detection:} While the vehicle is stationary, the state estimation exhibits random pose drift. This is a known problem in vehicle localization using inertial measurements \cite{vilens}. In this case, the state observability degrades dramatically due to insufficient IMU excitation, leading to unbounded error accumulation. Thus, we follow the idea proposed in \cite{vilens} to detect near-zero velocity motion by voting through multiple sensors that provide velocity information. If the vehicle is voted to be stationary, we temporally pause the graph optimization and state propagation.  
\end{remark}

\begin{remark} \label{rem: opt_freq}
     \textbf{Optimization Frequency:} By default, we extend and optimize the time-centric graph at $\SI{10}{Hz}$ in our experiments to achieve a good balance of accuracy and runtime efficiency. Although these frequencies can be flexibly configured in the proposed estimation framework, we found that optimizing the graph at $\SI{5}{Hz}$ is a threshold to avoid discontinuities (jumps) of the estimated trajectory in our application. For applications restricted by low-performance computing devices, choosing a higher frequency to extend the graph and a lower frequency for optimization can be considered.
\end{remark}

\subsection{Implementation}
We implemented our approach in \texttt{C++} using Robot Operating System \texttt{ROS2}\footnotemark[10]. The open-source software library \texttt{GTSAM}\footnotemark[11] was extended to implement the graph and factor formulations. We adopted the software solution for lidar odometry from \texttt{LIO-SAM}\footnotemark[12], where only the front-end feature extraction and association were adapted in our work. We used the positioning and orientation estimation solution from a dual-antenna GNSS setup to initialize the state variable $\myFrameVec{x}{0}{}$. In this work, we used a laptop with an Intel i9-9900K, \SI{16}{cores} at maximum $\SI{4.7}{G\Hz}$ and $\SI{64}{GB}$ memory for sensor pre-processing and graph optimization in experimental studies. 

\footnotetext[10]{\url{https://docs.ros.org/en/humble/index.html}}
\footnotetext[11]{\url{https://gtsam.org}}
\footnotetext[12]{\url{https://github.com/TixiaoShan/LIO-SAM}}


\section{Measurement Setup and Test Sequences}
\subsection{Measurement Setup}
\begin{figure}[!t]
    \centering
    \includegraphics[width=0.48\textwidth]{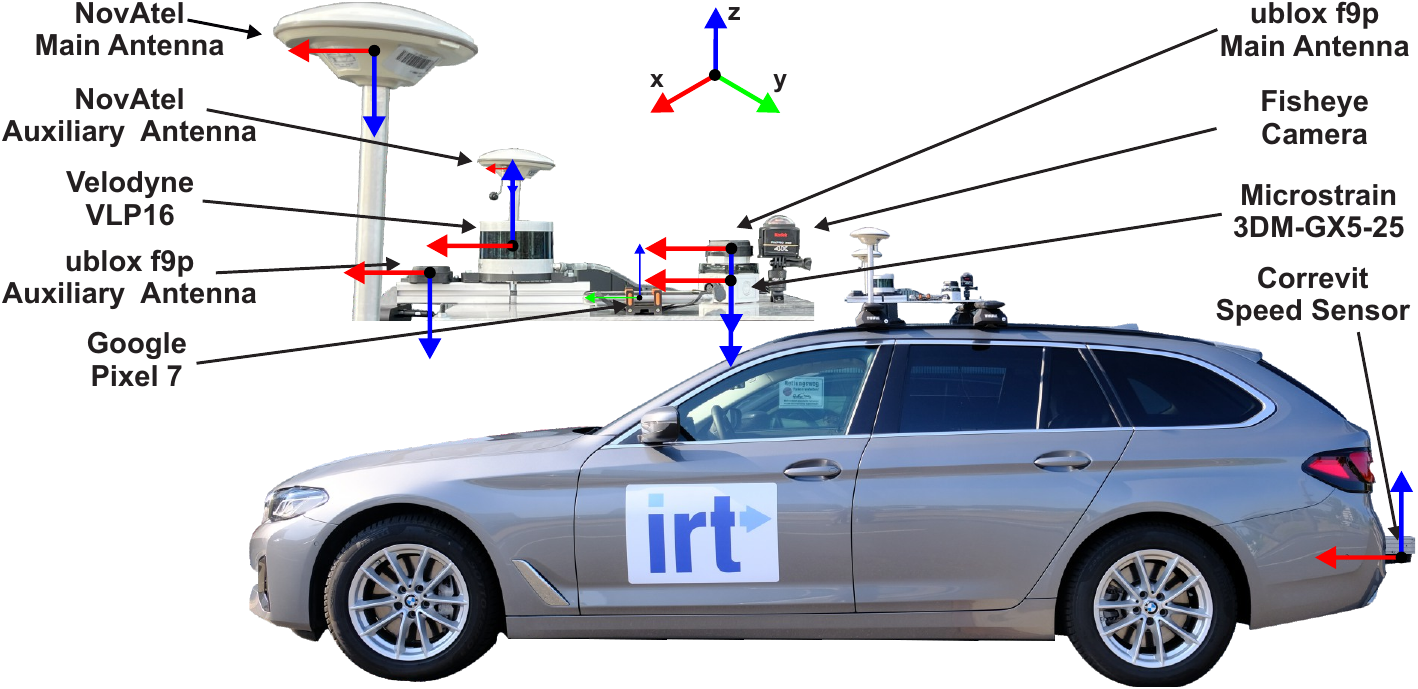}
    \caption{Sensor setup and frames on the test vehicle.}
    \label{fig: mea_hardware}
\end{figure}

In the measurement campaigns, we recorded sensor data of long-range routes in different areas of Aachen, Düsseldorf and Cologne. Our sensor setup included two GNSS receivers, a Microstrain 3DM-GX5 IMU, and a Velodyne VLP-16 lidar. Both the high-grade GNSS receiver (NovAtel PwrPak7D-E1) and the low-grade GNSS receiver (ublox f9p) were equipped with dual GNSS antennas and served with RTK correction data received from a base station. A high-grade optical speed sensor, Correvit S-Motion DTI from Kistler, was mounted on the trailer hitch on the vehicle's rear side. The sensor-equipped test vehicle is shown in Fig.\,\ref{fig: mea_hardware}. We have manually calibrated the static transformations between different sensors mounted on the roof rack of the test vehicle. Static transformation of the speed sensor to other sensors was measured using a Leica total station. These static transformations are assumed to be constant in all experiments. For more details, see our code\footnotemark[2].

The IMU data were acquired at $\SI{200}{\Hz}$, while lidar pointclouds were recorded at $\SI{10}{\Hz}$. GNSS observations from the NovAtel and ublox receiver were recorded at $\SI{10}{\Hz}$ and $\SI{5}{\Hz}$, respectively. We used the high-grade GNSS receiver Novatel PwrPark7D-E1 with a dual constellation of GPS and Galileo satellite systems as a reference source. In addition to the sensor data, we received the pulse-per-second signal (1PPS) from the NovAtel receiver at $\SI{1}{\Hz}$ to calculate the measurement delays. The $\mathrm{RTCMv3}$ (RTK) correction data from the German satellite positioning service \textit{SAPOS}\footnotemark[13] was also stored at about $\SI{1}{\Hz}$ for GNSS pre-processing.
\footnotetext[13]{\url{http://www.sapos.nrw.de}}

\subsection{Test Sequences} \label{sec: test_seq}
Our dataset contains different driving scenarios: open-sky, semi-/dense-urban, and high-speed track. For a clear evaluation, we define different test sequences throughout multiple measurement campaigns and analyze the driving conditions for each sequence, as shown in Table\,\ref{tab: test_sequences}. The test sequences include lengthy runs with a maximum $\SI{17}{\km}$ route, aiming to evaluate the estimation performance for long-term operations. For test sequences in urban areas, we chose data from scenarios with different urbanization rates containing tunnel and bridge crossings to evaluate the limitations of the proposed fusion approaches. In addition, we also considered open-sky areas on the high-speed track, where a maximum vehicle speed of $\SI{170}{\kilo\metre/\hour}$ was reached, creating significant motion distortion in the lidar point clouds. 

\begin{table}[!t]
\caption{Test sequences definition. We denote the test sequences in Aachen, Düsseldorf, Cologne, and high-speed tracks with ``AC", ``DUS", ``C", and ``HS", respectively. The variable $\myFrameScalarBar{v}{}{}$ represents the average speed and the scalar $\myFrameScalarBar{n}{}{\mathrm{sat.}}$ is the average number of satellites used for a GNSS-PVA solution. We calculated the ratio of RTK-fixed solution and No-Solution due to insufficient GNSS observations denoted by $\myFrameScalar{R}{\mathrm{fixed}}{\mathrm{RTK}}$ and $\myFrameScalar{R}{}{\mathrm{NS}}$, correspondingly.}
\label{tab: test_sequences}
\centering
\resizebox{0.48\textwidth}{!}{\begin{tabular}{cccccccc} 
\hline
\hline
Seq. & \begin{tabular}[c]{@{}c@{}}Leng.\\ (\si{\kilo\metre}) \end{tabular}  
     & \begin{tabular}[c]{@{}c@{}}Tunnel\\ (\si{\metre}) \end{tabular}  
     & \begin{tabular}[c]{@{}c@{}}Dura.\\ (\si{\second}) \end{tabular}  
     & \begin{tabular}[c]{@{}c@{}}$\myFrameScalarBar{v}{}{}$  \\ (\si{\kilo\metre/\hour}) \end{tabular}
     & $\myFrameScalarBar{n}{}{\mathrm{sat.}}$
     & \begin{tabular}[c]{@{}c@{}}$\myFrameScalar{R}{\mathrm{fixed}}{\mathrm{RTK}}$  \\ [0.1cm] (\si{\percent})\end{tabular}
     & \begin{tabular}[c]{@{}c@{}}$\myFrameScalar{R}{}{\mathrm{NS}}$  \\ [0.1cm] (\si{\percent})\end{tabular}\\
\hline
AC   & 17.0 & 270 & 2477 & 27.25  & 11 & 76.51 & 1.7  \\
\hline
DUS  & 5.25 &  -  & 1350 & 13.48  & 8  & 52.06 & 0.9  \\
\hline
C01  & 0.81 & 276 & 160  & 17.89  & 7  & 60.8  & 31.78 \\
C02  & 1.45 & 145 & 390  & 13.36  & 7  & 37.74 & 11.56 \\
\hline
HS  & 10.6 & -   & 300  & 124.82 & 14 & 94.9 & 1.47 \\
\hline\hline
\end{tabular}}
\vspace{-5mm}
\end{table}

\subsection{Reference Trajectory and Metrics}
To evaluate the proposed fusion strategies, we employ the RTK-fixed GNSS-PVA solution associated with low uncertainties ($\sigma_{\rm pos} < \SI{0.05}{m}$ and $\sigma_{\rm rot} < \SI{1}{\degree}$) to calculate the absolute root mean square error (RMSE). Besides the error metrics, we employ Pythagoras' theorem implemented in the Open Motion Planning Library\footnotemark[14] (OMPL) to calculate the trajectory smoothness (contrary to trajectory roughness) for all test sequences, aiming to provide a relative performance metrics. The smoothness is given as the sum of angles between all path segments in the local-world frame, as denoted in \eqref{eq: smoothness}, where the variables $a_i$, $b_i$ and $c_i$ are the length of the trajectory segments containing three successive vehicle positions in the Euclidean frame. For the same test sequence with $k$ vehicle positions, a smaller $s$ shows a high smoothness of the trajectory. In this work, we used the propagated states from the state publisher at a high frequency to calculate the smoothness, $s$:
\begin{equation}
   s = \sum_{i=2}^{k-1}\Bigg(\frac{2(\pi - \arccos{\frac{a_i^2+b_i^2-c_i^2}{2a_i b_i}})}{a_i+b_i}\Bigg)^2.
    \label{eq: smoothness}
\end{equation}
\footnotetext[14]{\url{https://ompl.kavrakilab.org/}}

\section{Experiments and Results} \label{sec: exp}

\subsection{Experiment Design}
To evaluate the proposed \texttt{gnssFGO}, we first benchmark the loosely coupled fusion of the GNSS solution with the IMU, 2-D speed sensor, and a lidar-centric SLAM approach \texttt{LIO-SAM} \cite{liosam}, aiming to evaluate the robustness of the proposed method. Compared to other multi-sensor fusion approaches, \texttt{LIO-SAM} represents a classical multi-sensor fusion framework performing well in outdoor scenarios where lidar odometry is the primary sensor. For a fair evaluation, we have adapted the \texttt{LIO-SAM} implementation\footnotemark[6] using the same robust error models and parameterizations as in our method. We also enable the loop-closure detection in \texttt{LIO-SAM} to maximize state estimation performance. We follow the implementation in \cite{liosam} to eliminate motion distortions in lidar points using the IMU measurements.

Furthermore, we evaluate two fusion mechanisms with different sensor modalities. In the loosely coupled fusion, we conduct fusion configurations of the IMU and the GNSS-PVT solution with and without multi-sensor including a 2-D speed sensor and lidar odometry (w. and w/o. MultiSensor). Later, we propose similar experiments by fusing raw GNSS observations and IMU measurements with and without lidar odometry in a tight coupling (w. and w/o. LiDAR), which is expected to present a more robust trajectory estimation in challenging areas compared to the loose coupling.

Lastly, we discuss the smoother type and computation time using different lag sizes. We also evaluate the GP-WNOJ prior and the GP-WNOA prior.

\makeTableMetricLC
\subsection{General Error Metrics}
With pre-defined test sequences in Table\,\ref{tab: test_sequences}, we present the general error metrics for all experiments in Table\,\ref{tab: err_metrics} by taking the RTK-fixed GNSS-PVA solution as the ground truth. Because an RTK-fixed solution is not available in challenging areas, we denote the solution rate used as a ground-truth reference to calculate the error metrics of each test sequence as a percentage in the column ``Seq.". Due to limited space, figures cannot be presented on a full scale; we thus upload all interactive figures in \texttt{GitHub}\footnotemark[15].
\footnotetext[15]{\url{https://github.com/rwth-irt/gnssFGO/tree/ros2/online_fgo/plots_tro}}

\begin{figure*}[h!]
\centering
\subfloat[Trajectories of Seq. AC.]{\includegraphics[width=0.46\textwidth]{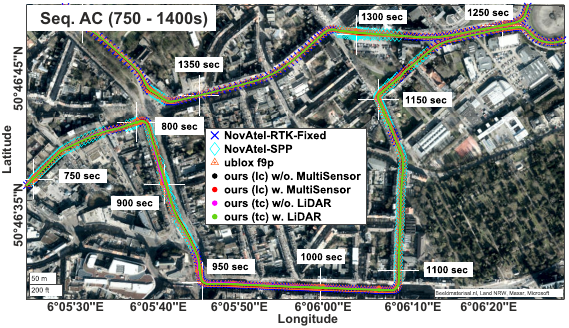}\label{fig: ac_map}}\hskip1ex
\subfloat[Coordinates in the WGS84 frame.]{\includegraphics[width=0.46\textwidth]{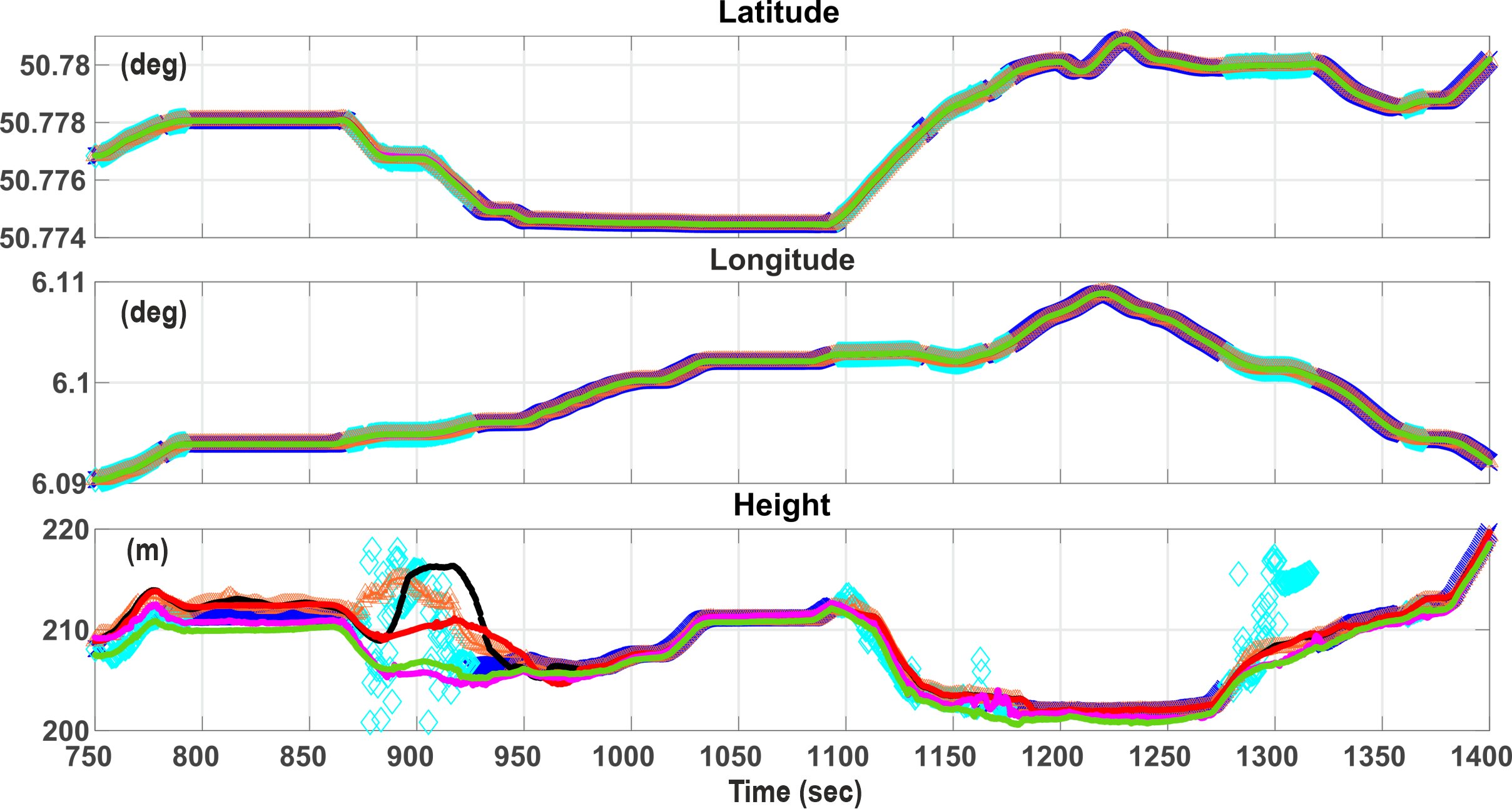}\label{fig: ac_pos}}\hskip1ex
\subfloat[Estimated rotation in the NED frame.]{\includegraphics[width=0.46\textwidth]{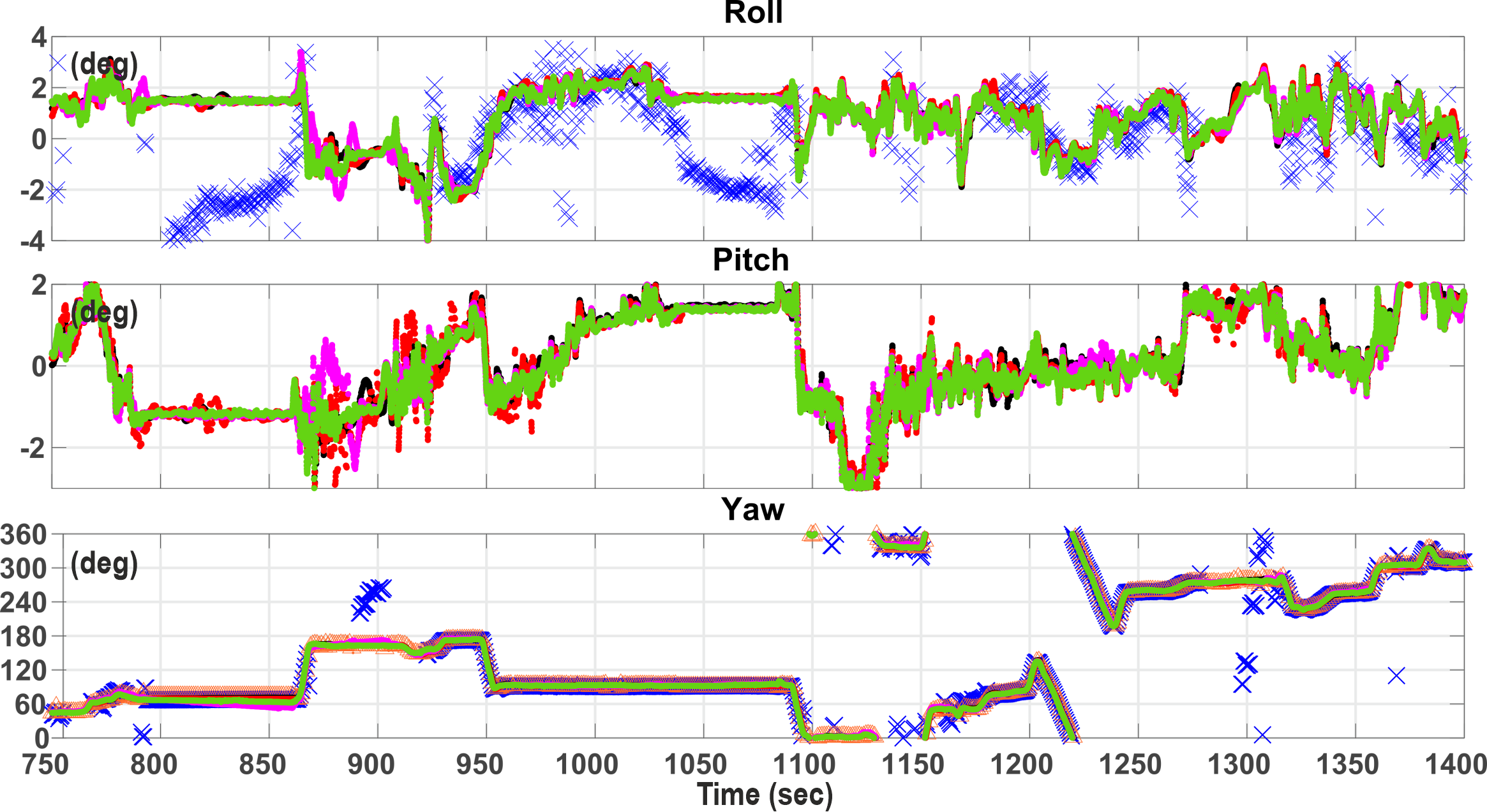}\label{fig: ac_rot}}\hskip1ex
\subfloat[Estimated velocity in the NED frame.]{\includegraphics[width=0.46\textwidth]{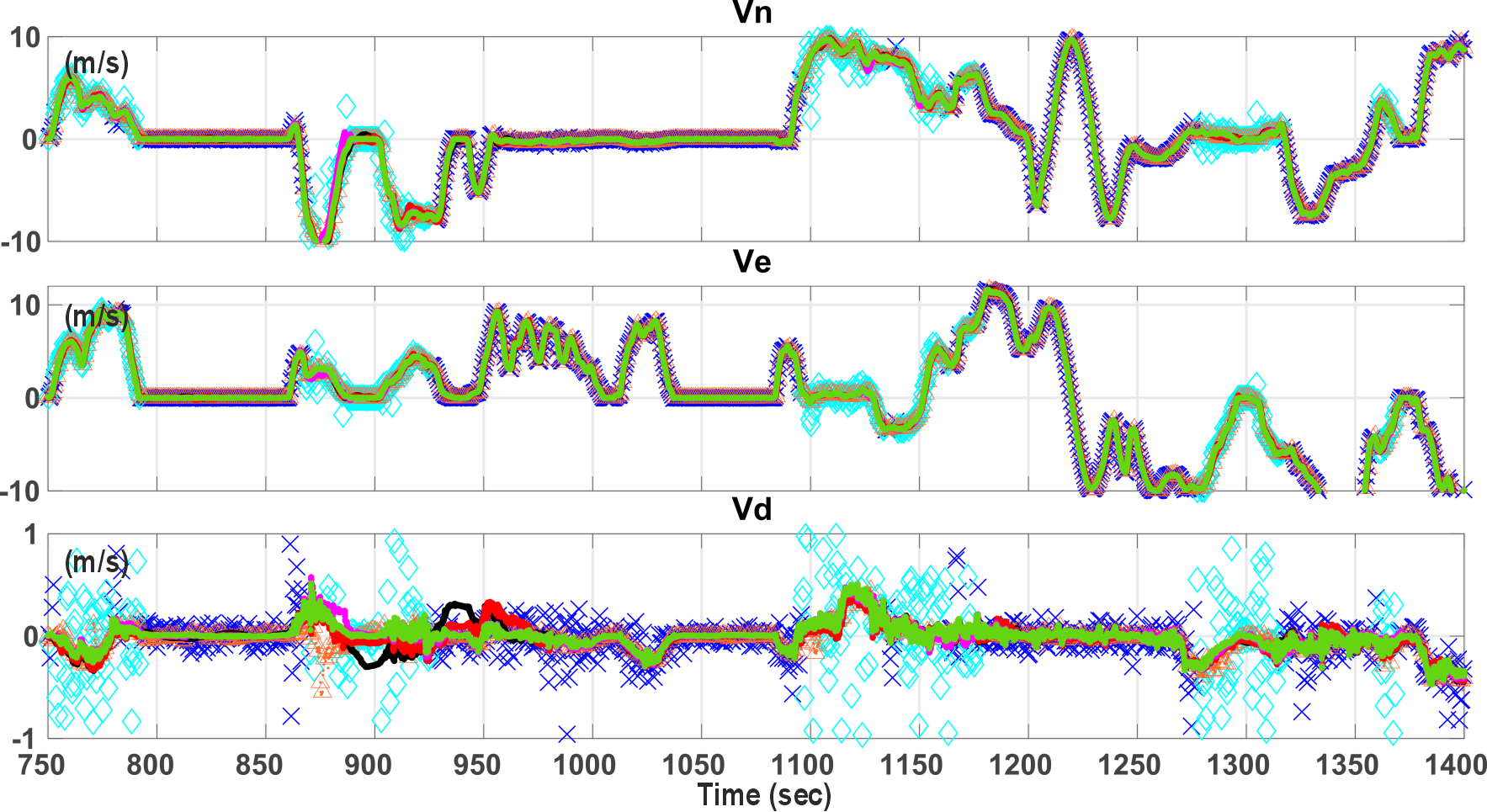}\label{fig: ac_vel}}\hskip1ex
\caption{Trajectory plot ($\SI{700}{s}$ - $\SI{1400}{s}$) in urban areas in Aachen. We plot the GNSS single point position (SPP) if the RTK-fixed solution is unavailable.}
\label{fig: traj_ac}
\end{figure*}

\begin{figure*}[h!]
\centering
\subfloat[Trajectories of Seq. DUS.]{\includegraphics[width=0.46\textwidth]{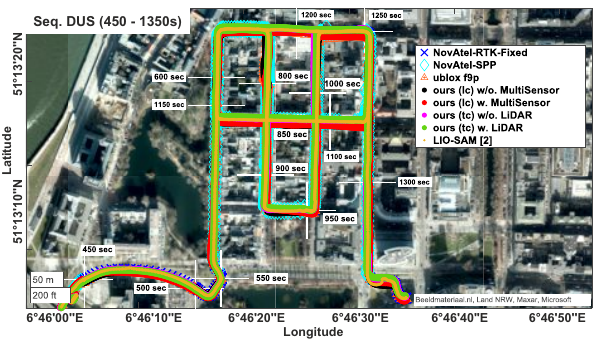}\label{fig: dus_map}}\hskip1ex
\subfloat[Coordinates in the WGS84 frame.]{\includegraphics[width=0.46\textwidth]{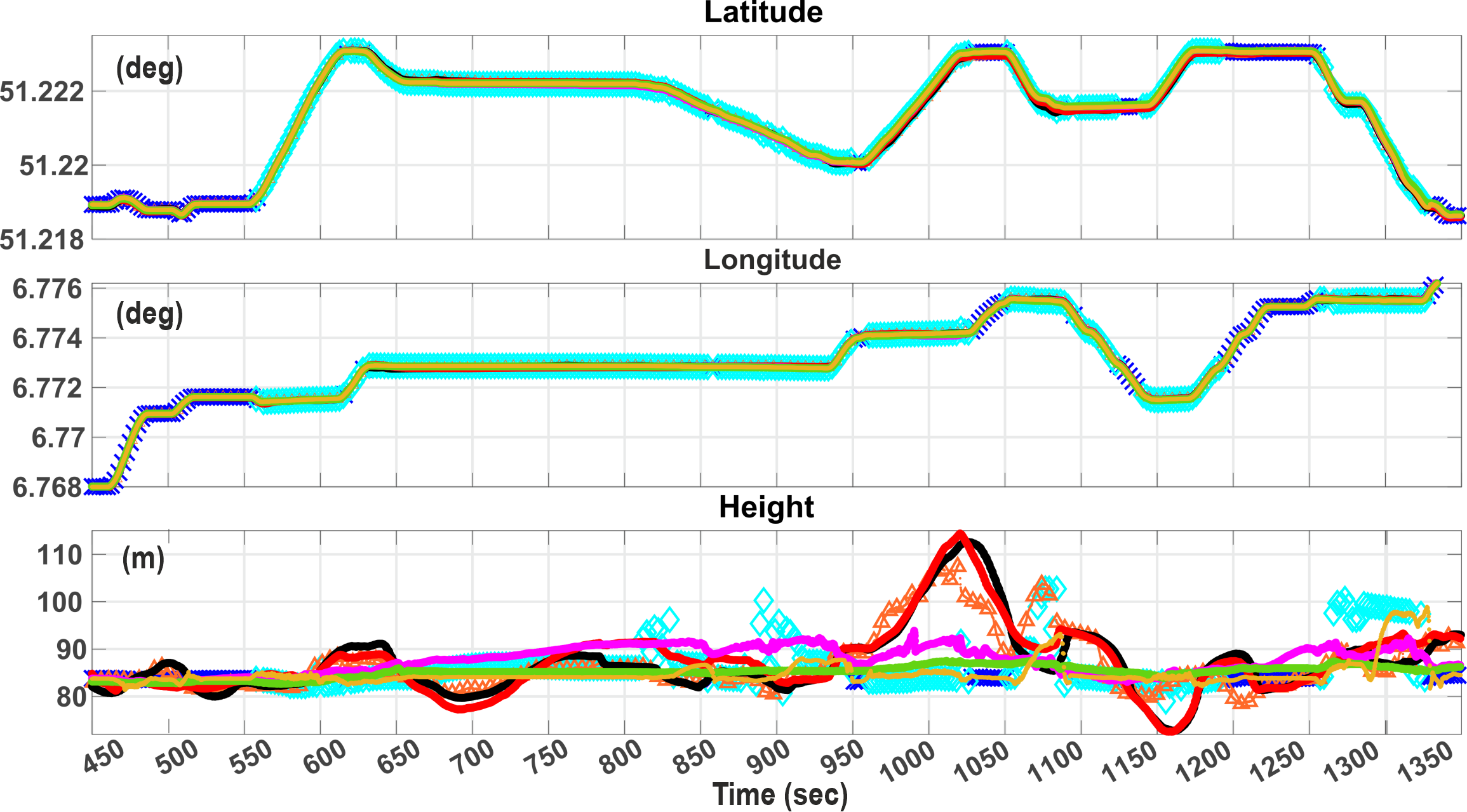}\label{fig: dus_pos}}\hskip1ex
\subfloat[Estimated rotation in the NED frame.]{\includegraphics[width=0.46\textwidth]{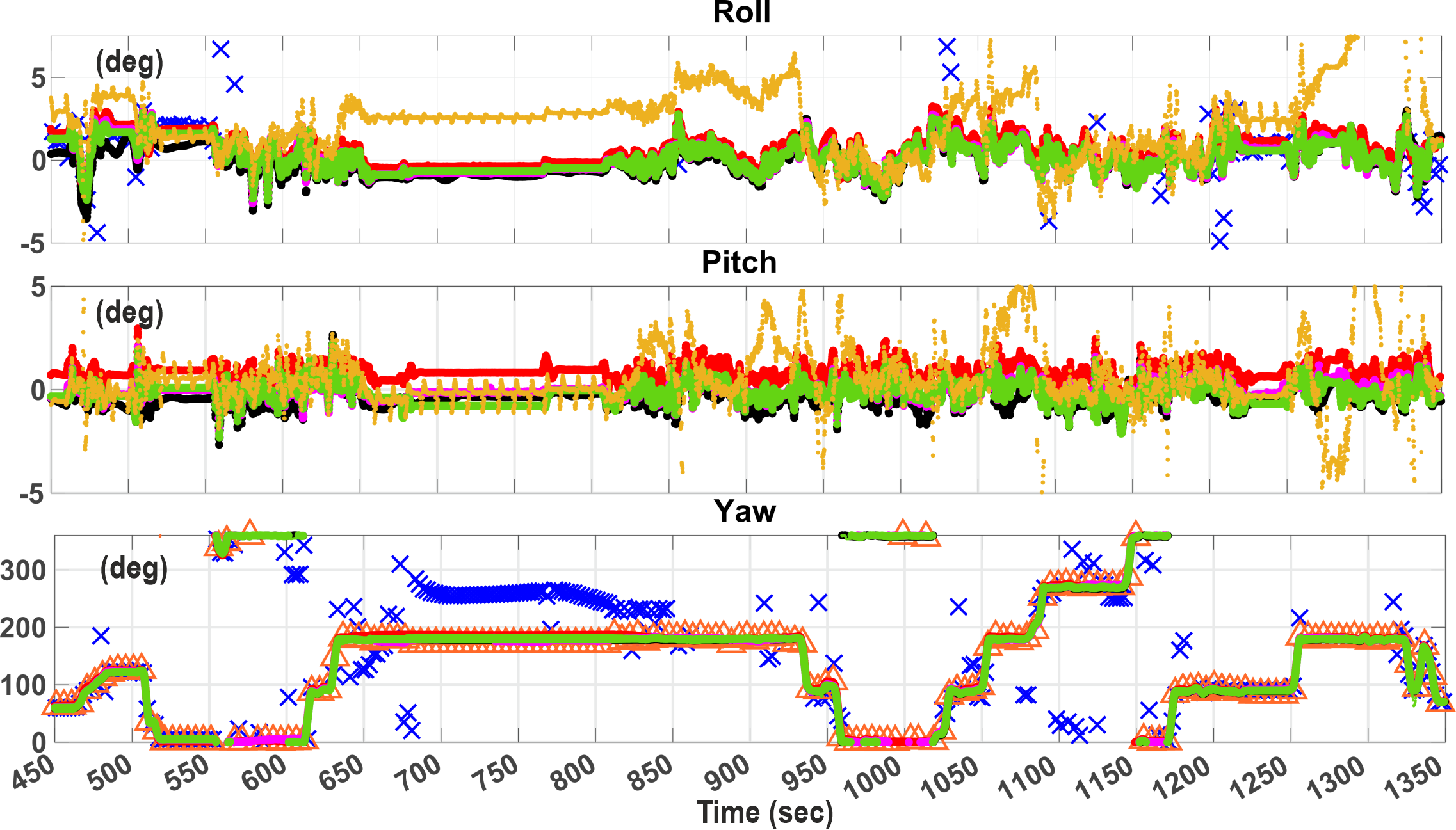}\label{fig: dus_rot}}\hskip1ex
\subfloat[Estimated velocity in the NED frame.]{\includegraphics[width=0.46\textwidth]{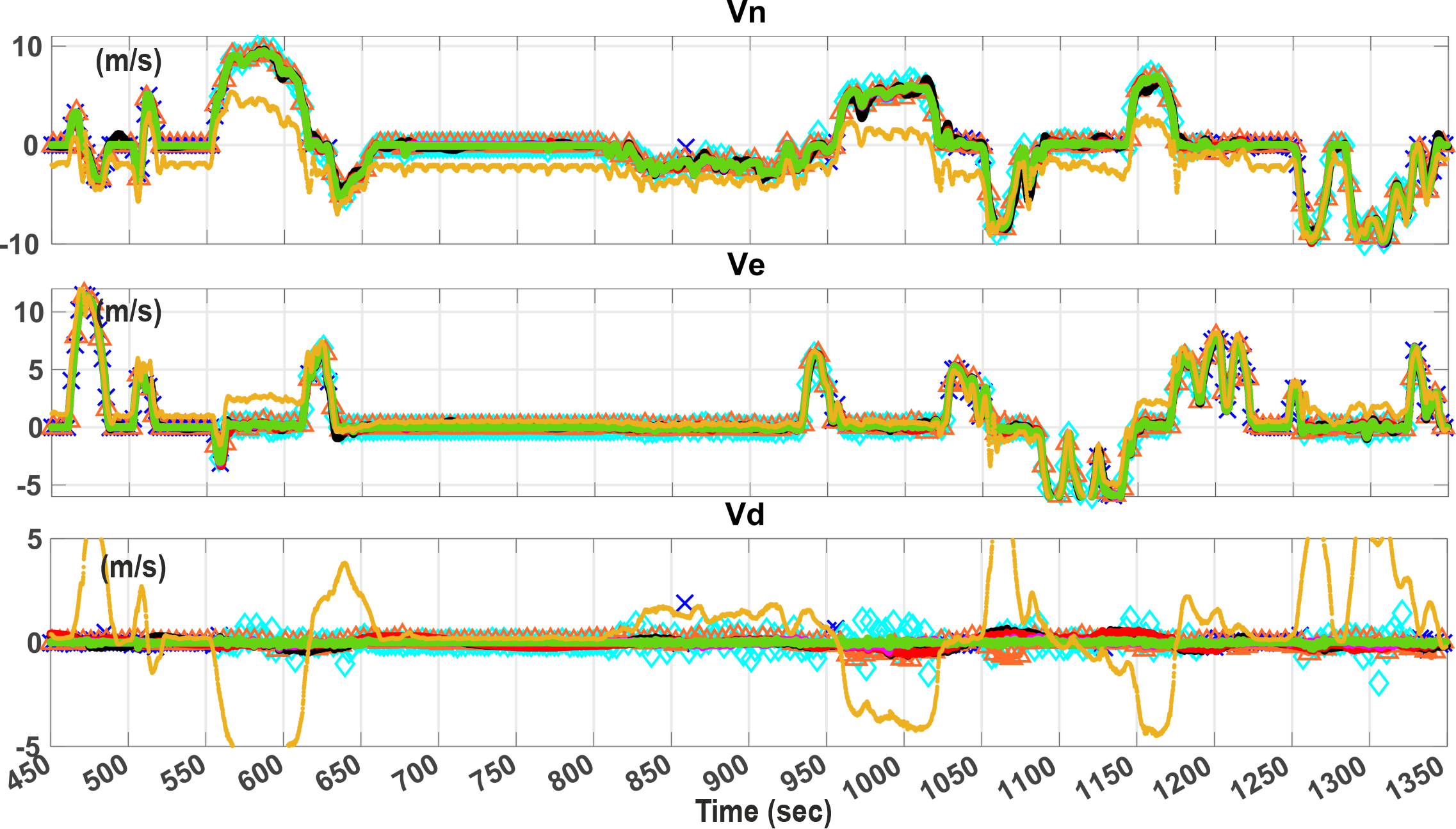}\label{fig: dus_vel}}\hskip1ex
\caption{Trajectory plot ($\SI{450}{s}$ - $\SI{1350}{s}$) in challenging areas in Düsseldorf.}
\label{fig: traj_dus}
\vspace{0.5cm}
\end{figure*}

\begin{figure*}[h!]
\centering
\subfloat[Trajectories of Seq. C01.]{\includegraphics[width=0.46\textwidth]{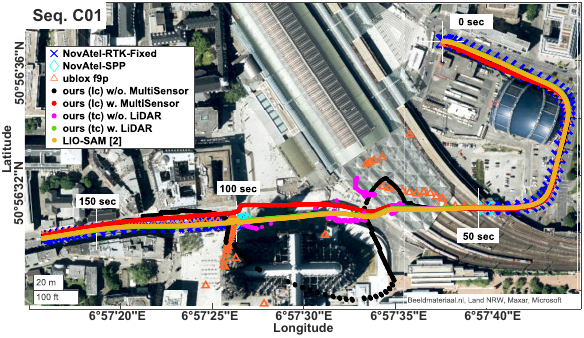}\label{fig: c01_map}}\hskip1ex
\subfloat[Coordinates in the WGS84 frame.]{\includegraphics[width=0.46\textwidth]{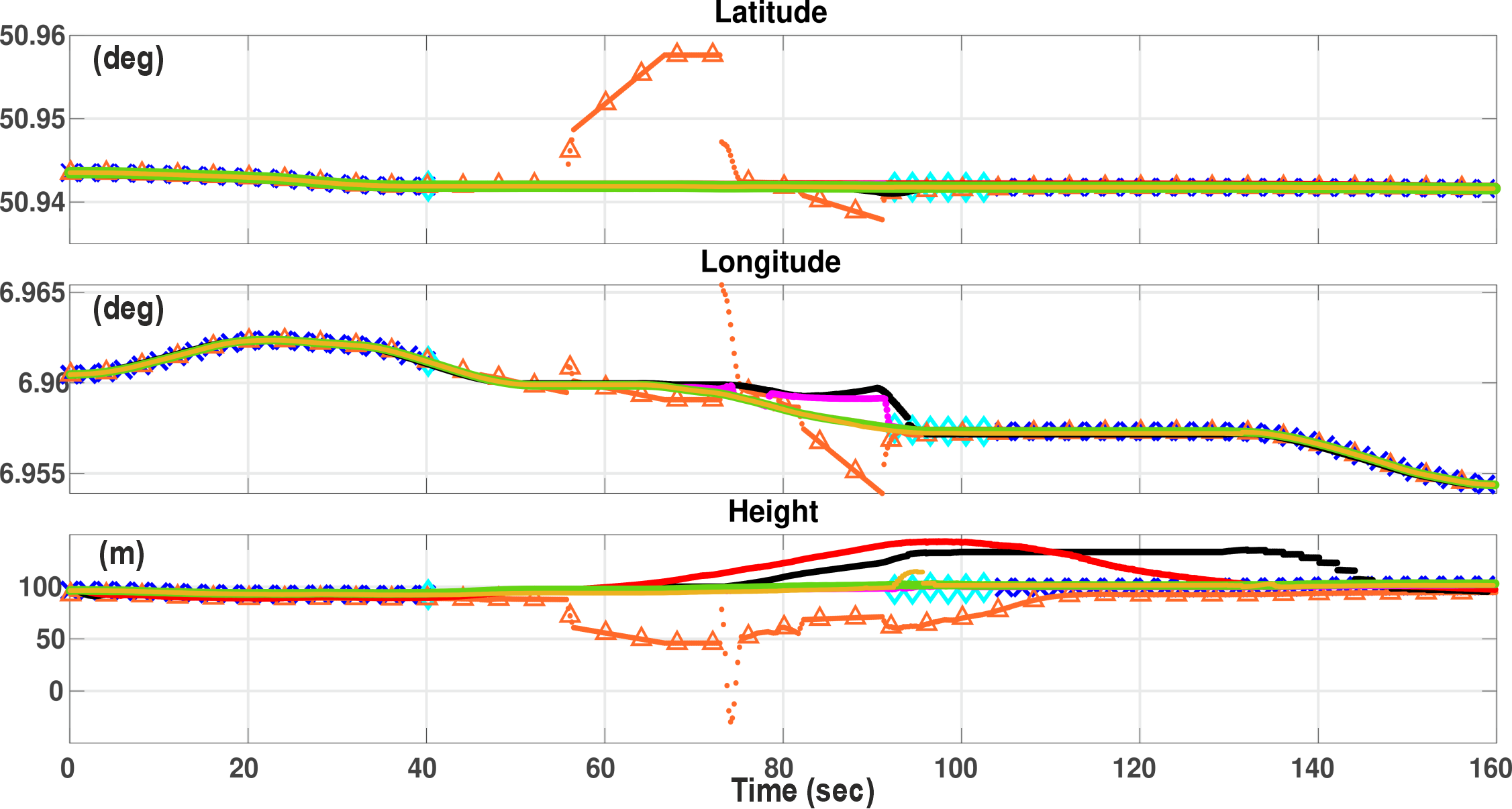}\label{fig: c01_pos}}\hskip1ex
\subfloat[Estimated rotation in the NED frame.]{\includegraphics[width=0.46\textwidth]{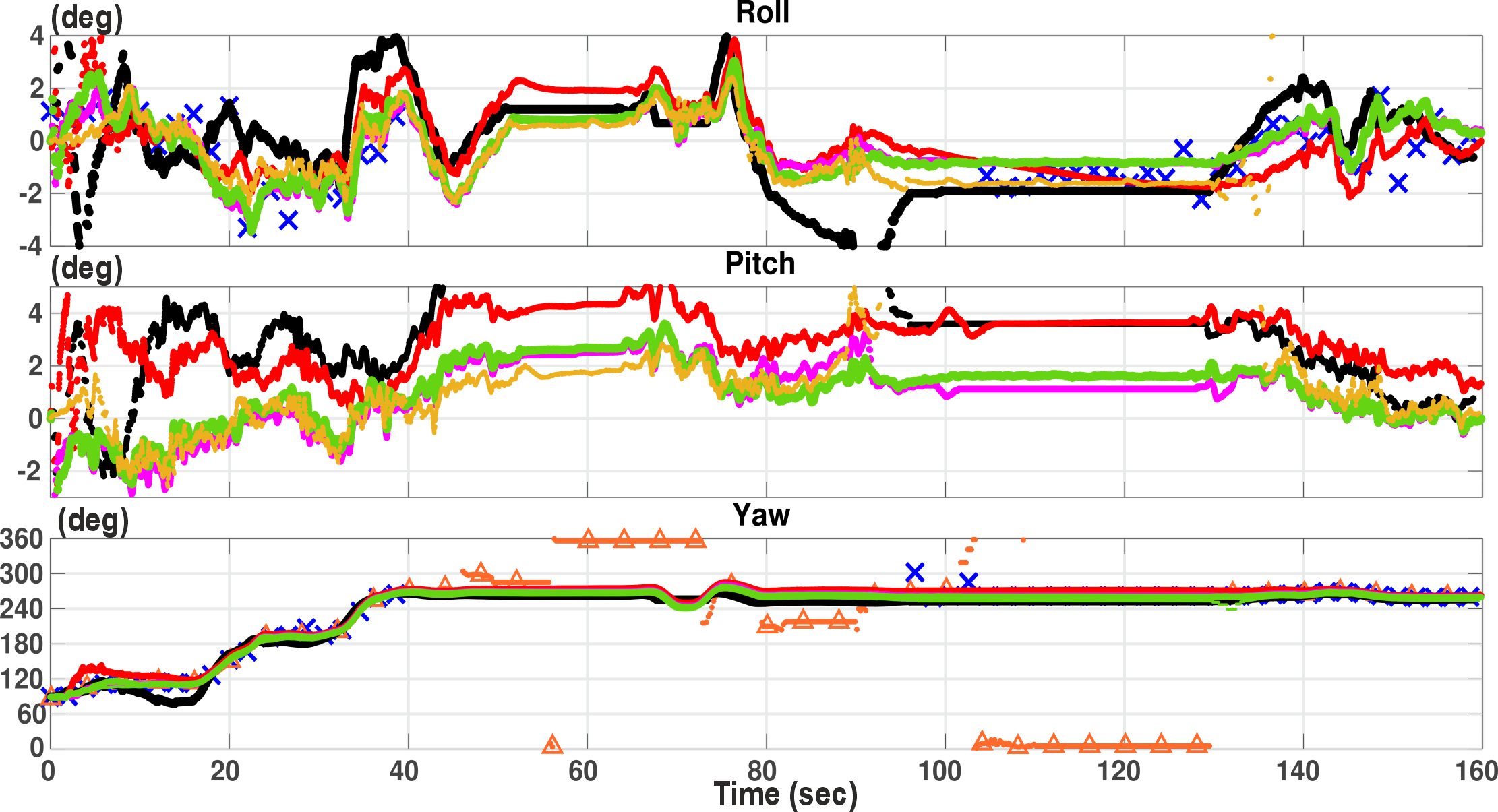}\label{fig: c01_rot}}\hskip1ex
\subfloat[Estimated velocity in the NED frame.]{\includegraphics[width=0.46\textwidth]{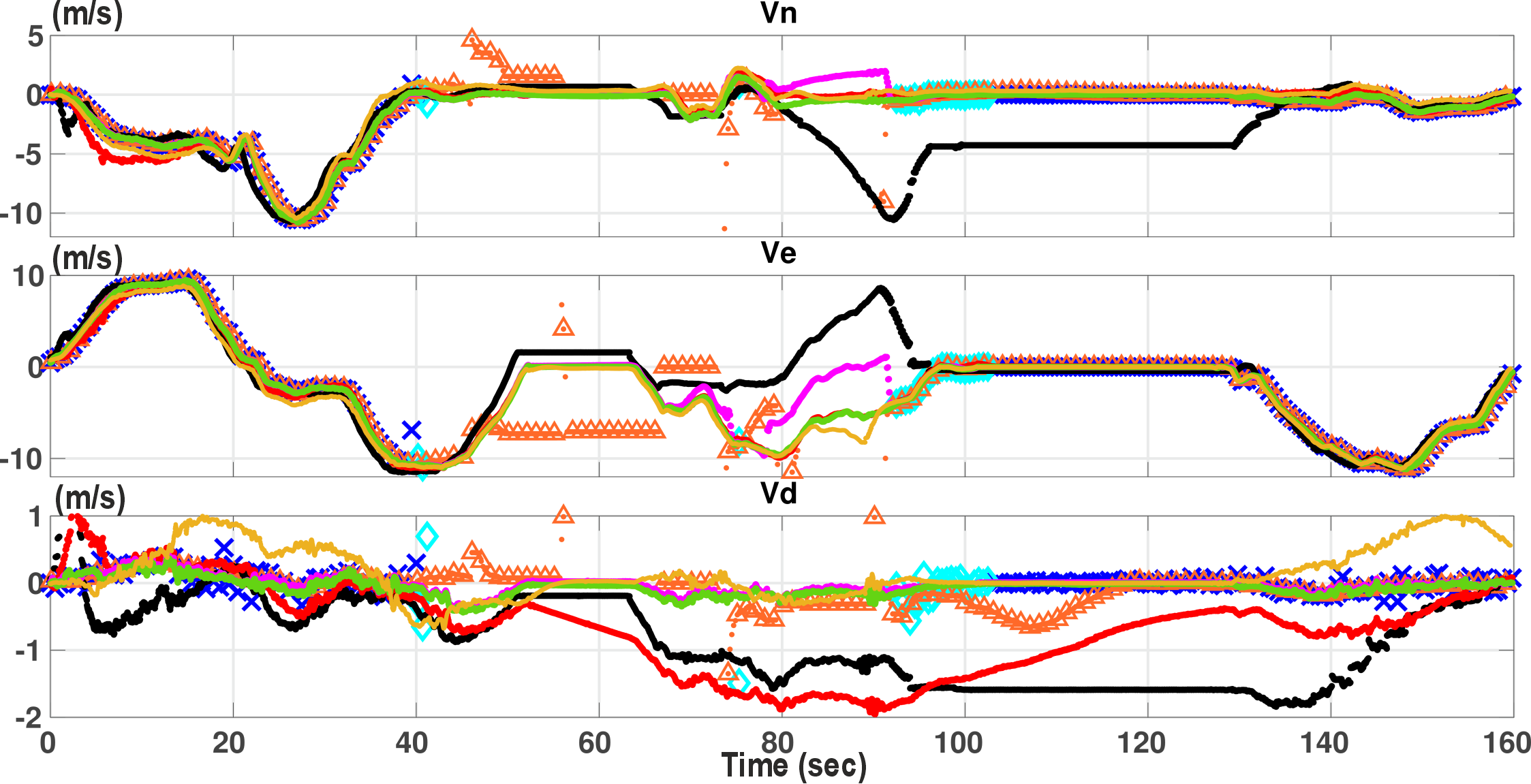}\label{fig: c01_vel}}\hskip1ex
\caption{Trajectory plot near the central station of Cologne (Seq. C01). For the tightly coupled fusion, we omitted the near-zero velocity detection to present trajectory drifting while the receiver clock error is unobservable.}
\label{fig: traj_c01}
\end{figure*}

\subsubsection{Lidar-Centric Fusion}
As shown in Table\,\ref{tab: err_metrics}, the lidar-centric SLAM approach \texttt{LIO-SAM} failed in several test sequences even when the same factors with robust error modeling were used and loop-closure detection was enabled (see video demonstration\footnotemark[2]). The most frequent reason is that the scan registration fails due to an invalid feature association, which can be observed in all failed test sequences. Another possible reason for the failure can be associated with corrupted GNSS observations that show inconsistent noise values, resulting in a divergence in optimization. Fig.\,\ref{fig: c02_liosam} demonstrates this result, where the estimate diverged and cannot be recovered after the vehicle entered a tunnel. In Seq. HS, the lidar-centric approach cannot even be properly initialized while the vehicle moves very fast, which was not observed in the proposed \texttt{gnssFGO}. Furthermore, the estimated velocities and orientations using \texttt{LIO-SAM} show large variation and therefore less robustness compared to both fusion mechanisms in \texttt{gnssFGO} (see Fig.\,\ref{fig: dus_rot} and Fig.\,\ref{fig: dus_vel}). 

\textit{Discussion:} 
We observe that because graph construction triggered by the lidar odometer in \cite{liosam} requires strict timestamp synchronization of GNSS measurements with lidar timestamps, asynchronous GNSS measurements are dropped. Although lidar odometry and detected loop closures are still available,  dropping GNSS measurements results in a loss of effective state constraints, so graph optimization becomes more sensitive to inaccurate scan registration. Therefore, the trajectory's smoothness and the estimate's accuracy are dramatically penalized (see Table\,\ref{tab: err_metrics}). This hypothesis is supported in the test sequences DUS and C01 (see Fig.\,\ref{fig: traj_dus} and Fig.\,\ref{fig: traj_c01}), where the estimated height, orientation and velocities were frequently diverted. Therefore, it can be observed that trajectory drift cannot be effectively eliminated using the classic sensor-centric localization approach \texttt{LIO-SAM}. Even worse, the robustness and reliability of sensor-centric approaches cannot be guaranteed in challenging areas once the primary sensor is compromised. As online applications raise computation time and resource requirements, sensor degradation due to, e.g., insufficient data processing becomes non-trivial. The proposed \texttt{gnssFGO} presents an effective workaround while fusing multiple sensors to eliminate the dependence on a single sensor, enabling the fusion of lossless information and improving the robustness of the estimate if sensor failure can be expected.

\subsubsection{Loosely Coupled (lc) Fusion} \label{sec: res_lc}
It can be observed in Table\,\ref{tab: err_metrics} that the accuracy of the low-grade GNSS receiver, especially in the vertical dimension (height), is dramatically degraded compared to the high-grade GNSS receiver. Furthermore, this receiver does not characterize its noise values, so the standard deviations provided are strongly inconsistent with the real noise. Therefore, fusing the PVT solution from the low-grade GNSS receiver significantly downgrades the performance of the loosely coupled fusion in all test sequences, even when multiple sensor observations are fused. In a less challenging environment (Seq. AC), the proposed loosely coupled fusion generally outperforms the original PVT solution. However, the same improvement cannot always be expected in challenging environments when comparing the error metrics of other test sequences. The primary reason for this result is caused by inconsistent noise values and highly inaccurate height measurements in the PVT observations. Another interesting phenomenon is that the loosely coupled fusion of the PVT solution with other sensors can present a degraded performance (see Seq. C01 and Seq. HS). Due to inaccurate height measurements that present high variations (see Fig.\,\ref{fig: dus_pos} and Fig.\,\ref{fig: c01_pos}), the pose of the lidar keyframes cannot be stably optimized, so the keyframes, which are used to calculate the relative motion of each scan, present different height values. In such cases, the estimated lidar odometry is associated with incorrect relative motion increments, which degrades graph optimization (see height in Fig.\,\ref{fig: c01_pos} and Vd in Fig.\,\ref{fig: c01_vel}).

To improve the robustness of the estimation by acquiring redundant state constraints, we used the high-grade speed sensor in this fusion approach, expecting it to effectively constrain the unobserved states once the GNSS solutions are compromised. However, our experiments indicate that the 2-D velocity measurements provided by the 2-D speed sensor cannot sufficiently constrain the state space, especially when vehicle orientation cannot be observed.

\textit{Discussion:} Although the loosely coupled fusion using the proposed \texttt{gnssFGO} did not fail in our test sequences, we can observe the same conclusion as shown in \cite{ekf_fgo_compare} that loosely coupled fusion cannot serve as a robust state estimator in challenging areas where GNSS solutions present inconsistent uncertainties. However, we indicate that this fusion mechanism can present fast estimation convergence as long as accurate GNSS-PVT solutions are available. This is because the state variables (position and velocity) can be directly observed in the GNSS-PVT solution, where the measurement model does not present high nonlinearity, and thus effective state constraints are presented.

\subsubsection{Tightly Coupled (tc) Fusion}\label{sec: res_tc}
Compared to loosely coupled sensor fusion, fusing pre-processed GNSS observations in a tight coupling contributes to redundant state constraints. Thus, the tightly coupled fusion can generally present more robust trajectory estimations with a smaller maximum position error and larger trajectory smoothness in lengthy runs, except in the high-speed scenario (Seq. HS). 
In challenging urban areas such as Seq. C01 and C02, fusing the lidar odometry as between-state constraints generally improves the estimation performance and trajectory smoothness. This conclusion can also be drawn when referring to Fig.\,\ref{fig: dus_pos} and Fig.\,\ref{fig: c01_pos}, where more accurate height and velocity estimations can be observed by fusing lidar odometry in the graph. In high-speed scenarios, lidar scans suffer from serious motion distortion, and no sufficient features can be extracted compared to urban areas. Therefore, a limited performance improvement can be observed by fusing lidar odometry in the graph.

\textit{Discussion:} Based on the experimental results presented above, a robust trajectory estimation can be achieved in challenging scenarios using the proposed approach by fusing multiple sensor measurements in a tight coupling, which supports our hypothesis proposed in Sec.\,\ref{sec: rw}. In contrast to the loose coupling, the tightly coupled multi-sensor fusion presents a more robust trajectory estimation in our experimental studies. The same conclusion has also been shown in \cite{ekf_fgo_compare, gvins}. However, acceptable accuracy cannot be achieved, especially in dense urban scenarios. For instance, although all estimated trajectories using the proposed \texttt{gnssFGO} in Fig.\,\ref{fig: demo_liosam_tcfgo} remain consistent, a large drift is presented using the proposed sensor integration. Possible reasons to explain this phenomenon can be traced back to lidar degradation, insufficient GNSS observations, and inconsistent sensor noise models due to the presence of outliers. As \texttt{gnssFGO} provides a flexible fusion mechanism, this problem can be addressed by integrating more effective state constraints into the graph.

\subsection{Challenging Scenarios}
In this part, we propose experimental studies regarding GNSS observations, lidar odometry, and solver settings. We also evaluated the GP-WNOA and GP-WNOJ priors and discussed the hyper-parameter tuning for $\myFrameVec{Q}{c}{}$. 

\begin{figure}[!t]
    \centering
    \includegraphics[width=0.45\textwidth]{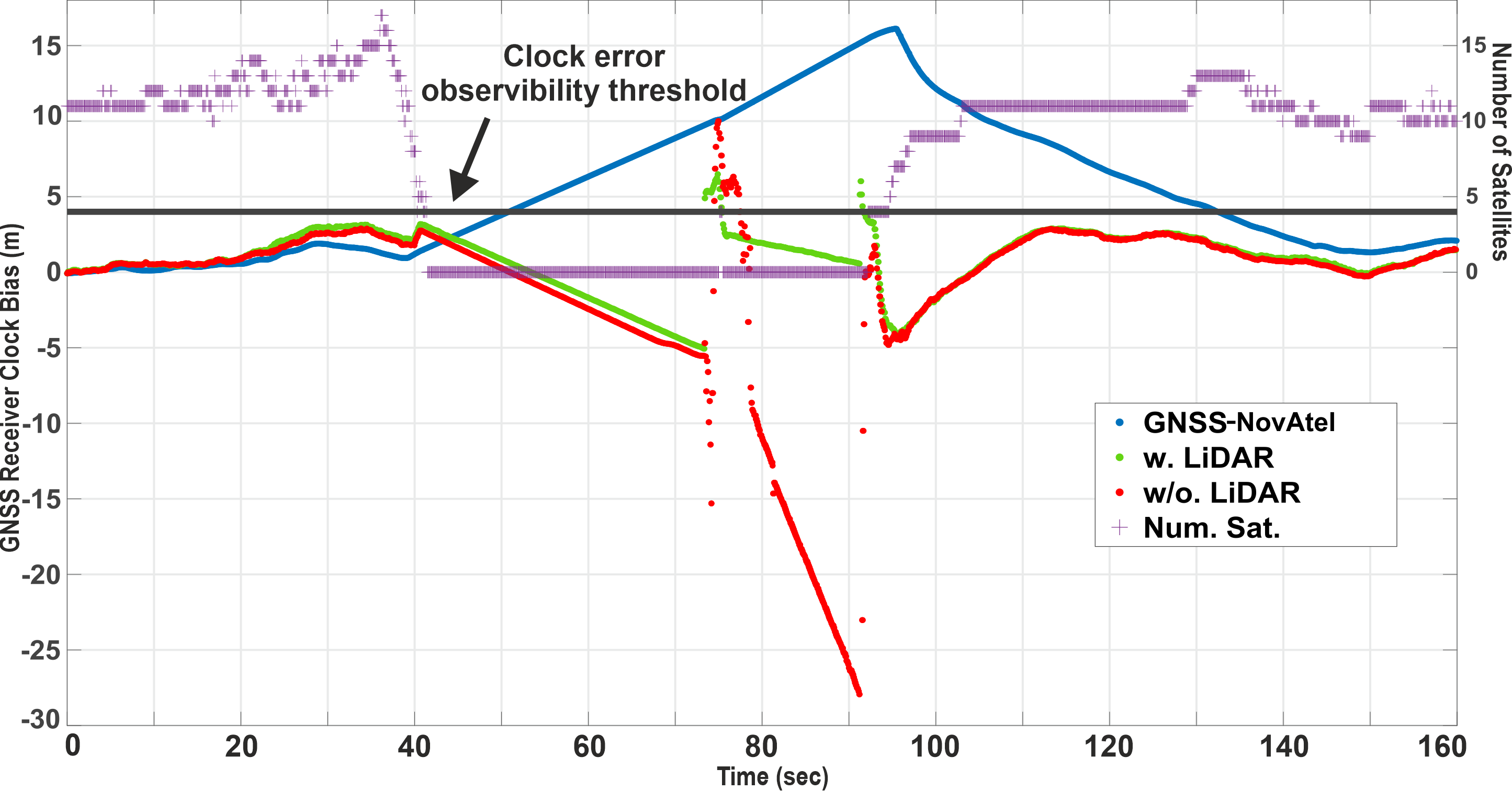}
    \caption{Estimated GNSS receiver clock bias of Seq. C01 without near-zero-velocity detection. The receiver clock error is unobservable during tunnel crossing. In this scenario, fusing raw GNSS observations without lidar odometry cannot constrain the state estimation, resulting in trajectory drifting. The corresponding trajectories of both fusion approaches are presented in Fig.\,\ref{fig: traj_c01}.}
    \label{fig: c01_clk}
\end{figure}

\subsubsection{Loss of GNSS Observation}
Generally, losing GNSS observations in a short time interval does not lead to immediate divergence or trajectory drift if multiple state constraints such as lidar odometry or motion prior factors are still presented. This conclusion can be drawn from our experiment in Seq. C01, where the vehicle crossed a large bridge at the central train station in Cologne, as shown in Fig.\,\ref{fig: traj_c01}. It is also interesting to observe that the loss of GNSS observations is frequently accompanied by highly corrupted GNSS observations due to multipath effects. For example, fusion approaches can diverge when the vehicle approaches or leaves a tunnel. In this case, robustness can be significantly affected even if no local sensors are fused.

Furthermore, fusing GNSS observations in a tight coupling extends the state variables with receiver clock bias and drift $\myFrameVec{c}{}{r}=[\myFrameScalar{c}{b}{}~\myFrameScalar{c}{d}{}]^T$, which become unobservable if less than four satellites are visible. Fig.\,\ref{fig: c01_clk} shows the estimated clock bias $\myFrameScalar{c}{b}{}$ with respect to the number of tracked satellites. The estimated clock bias drifts dramatically in the graph where only GNSS observations and IMU measurements are integrated. Even if the observability of the clock bias can be recovered, it takes some time until the state variable $\myFrameVec{c}{}{r}$ converges (see Fig.,\ref{fig: c01_clk}), which downgrades the overall estimation performance. Similar results can be observed in other experiments where unobservable state variables can lead to estimation divergence and an ill-posed optimization problem. Fortunately, this problem can be eliminated in the graph fused with lidar odometry. Thanks to the between-state constraints that prevent other state variables (e.g., position and rotation) from divergence, robust trajectory estimation can be guaranteed. In addition, a large trajectory drift can be expected if the global reference (e.g., GNSS observations) is lost over a long time interval, such as when crossing a long tunnel.

\subsubsection{Highly Corrupted GNSS Observations}
\begin{figure}[!t]
    \centering
    \includegraphics[width=0.48\textwidth]{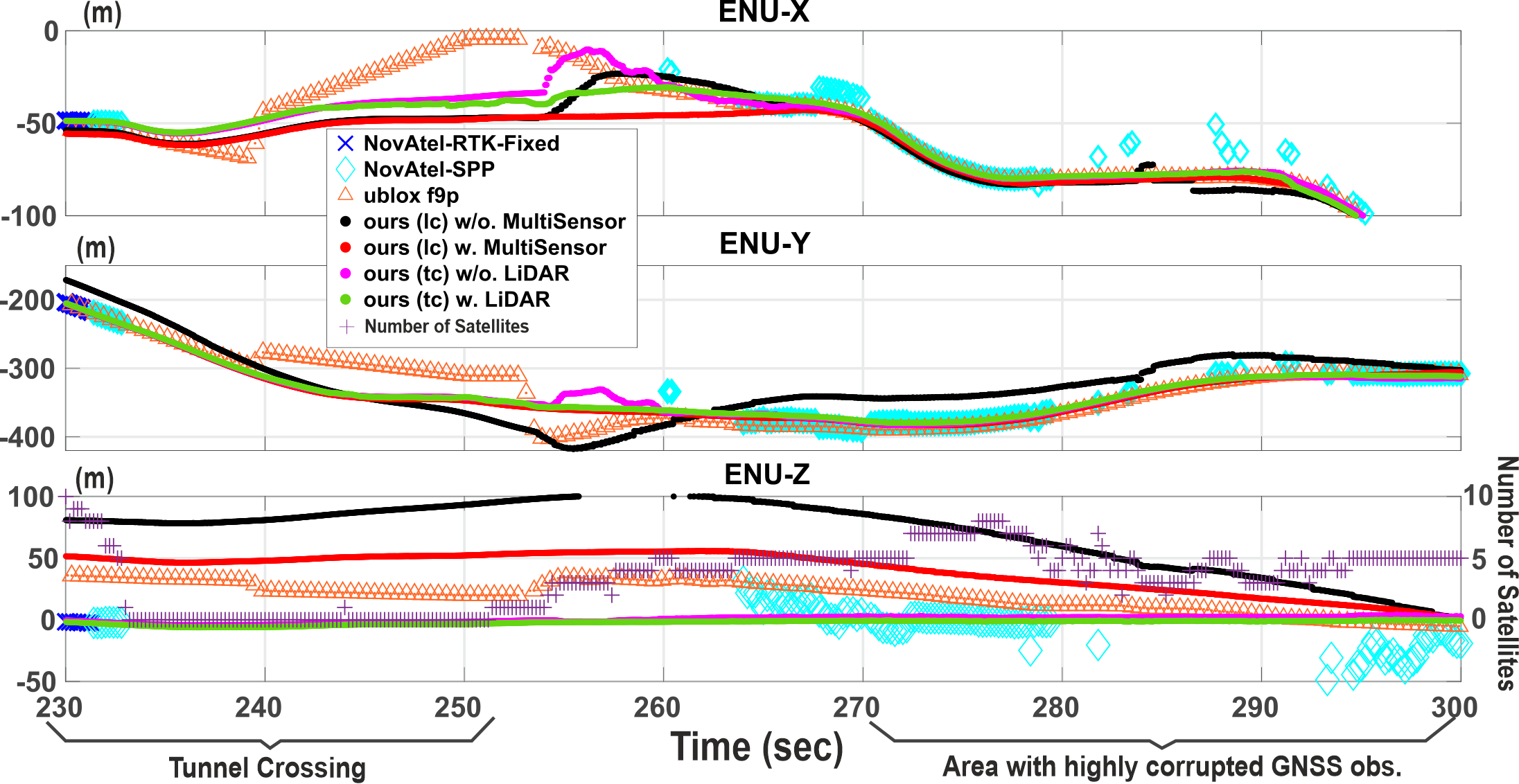}
    \caption{Coordinates in the ENU frame of Seq. C02. The estimated trajectories become unsmooth if the GNSS observations are strongly corrupted in urban areas.}
    \label{fig: c02_pos_enu}
\end{figure}
Compared to the temporary loss of GNSS observations, we emphasize that including highly corrupted GNSS observations in the graph has a greater impact on estimation performance. This conclusion can be supported by Seq. C02, where the accuracy of our proposed fusion paradigms is significantly degraded in GNSS-corrupted areas, as shown in Fig.\,\ref{fig: demo_liosam_tcfgo}. In Fig.\,\ref{fig: c02_pos_enu}, we plot the estimated trajectories in this scenario by transforming the coordinates in the navigation frame (ENU). A large trajectory drift up to $\SI{25}{\m}$ can be observed in tightly coupled fusion without lidar odometry (see Fig.\,\ref{fig: c02_pos_enu}). Although fusing relative motion constraints, such as odometry, can effectively constrain divergence, trajectory drifts cannot be eliminated until valid global references are acquired.

\subsubsection{Lidar Odometry Degradation}
\begin{figure}[!t]
    \centering
    \includegraphics[width=0.48\textwidth]{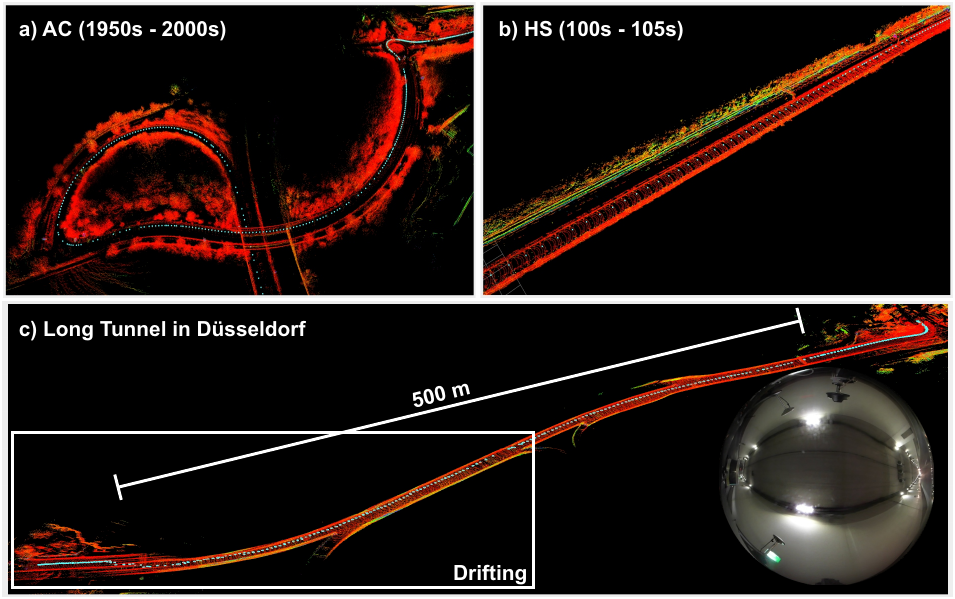}
    \caption{Examples of lidar odometry degradation in three scenarios: a) unstructured feature-less area, b) high-speed scenario, and c) long tunnel ($\SI{400}{\m}$).}
    \label{fig: lidar_degradation}
\end{figure}

As discussed in \cite{david_iros23}, traditional lidar odometry algorithms suffer from dramatic degradation in unstructured environments and high-speed scenarios. This problem can also be observed in our experiments. Fig.\,\ref{fig: lidar_degradation} illustrates three scenarios in which the accuracy of the lidar odometry is penalized if the vehicle is driving in featureless areas or in high-speed mode with an average vehicle speed of $\SI{125}{\kilo\metre/\hour}$. In low-speed driving mode and open-sky areas, lidar degradation does not reduce the estimation performance, while high-quality GNSS measurements are available. However, if the vehicle moves at high speed, the lidar odometry becomes inaccurate because of motion distortion. Therefore, including lidar odometry factors in the graph can decrease localization accuracy, as Table\,\ref{tab: err_metrics} of Seq presents. HS. In scenarios with long tunnels, trajectory drifting can always be expected due to the loss of global reference. This presents the major limitation of classic lidar odometers that calculate only pose increments. 
Another crucial aspect to be mentioned is the uncertainty of lidar odometry used in fusion approaches. Because no covariance is provided by classic scan matching algorithms, the pre-defined noise parameters (see Sec.\,\ref{sec: noise_model}) may be over-confident. Therefore, we emphasize the importance of acquiring realistic uncertainty quantification for lidar odometry in future work.

\subsection{Smoother Type and Computation Time} \label{sec: res_smoothers}
To study the impact of different smoother types and lag sizes, we evaluated batch and incremental smoother \texttt{iSAM2} with different lag sizes for Seq. DUS. The performance metrics are presented in Table\,\ref{tab: dus_sovler}. Compared to an incremental smoother, solving the optimization problem with a batch optimizer does not show a considerable improvement in accuracy. This happens because the graph structure becomes more similar to a Markov chain in large-scale localization applications with fewer loop-closure constraints. In this scenario, re-linearizing all past state variables does not contribute more information that improves the accuracy. For loosely coupled fusion, the batch smoother presents a smoother trajectory. However, this advantage is absent with the incremental smoother when fusing GNSS observations in a tight coupling. 

\begin{figure}[!t]
    \centering
    \includegraphics[width=0.48\textwidth]{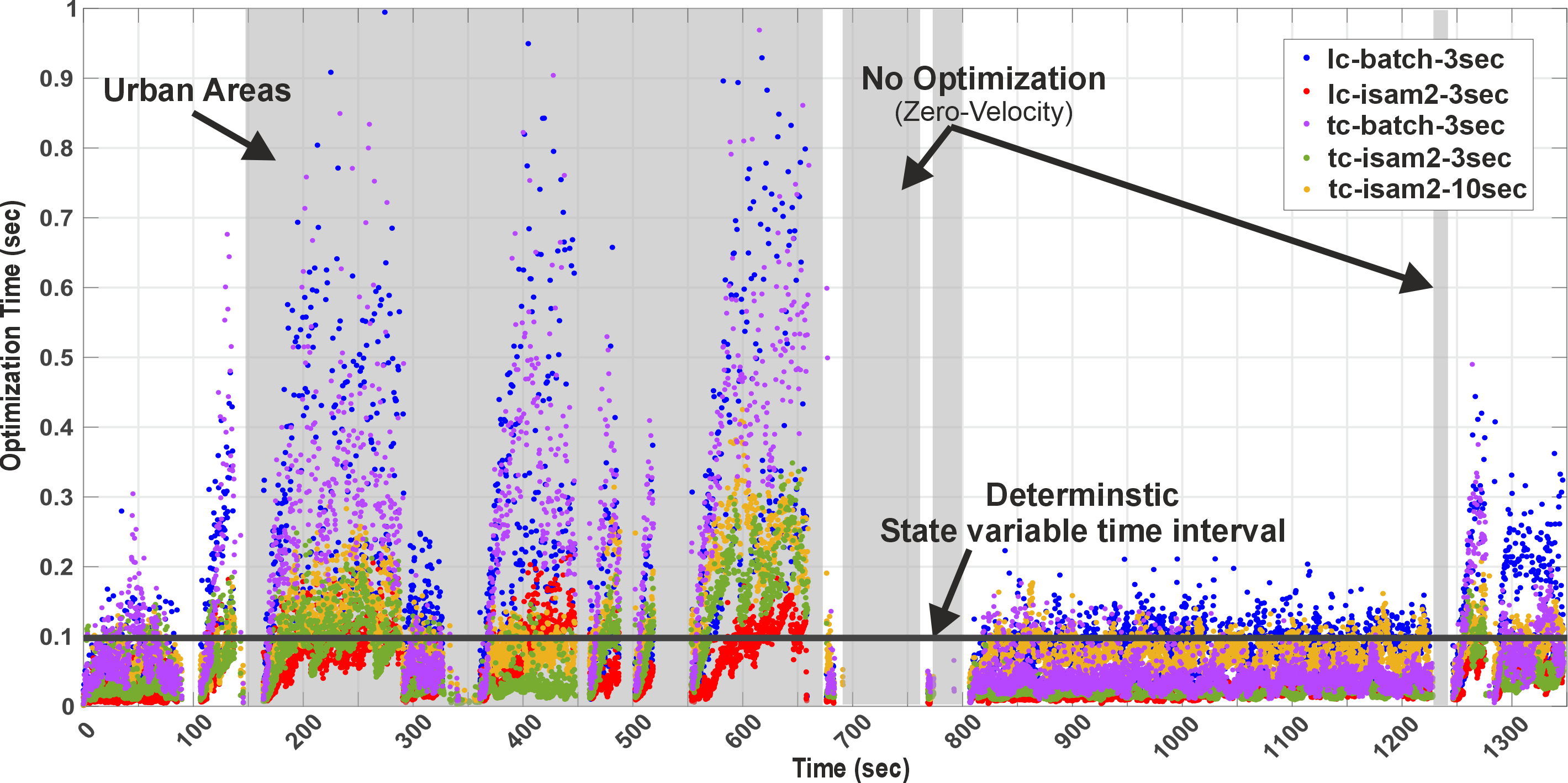}
    \caption{Computation time with different configurations of Seq. DUS.}
    \label{fig: dus_comp_time}
\end{figure}

Furthermore, the batch optimizer requires more computational resources than the incremental smoother (see Fig.\,\ref{fig: dus_comp_time}), especially in urban areas with more measurement outliers. In online applications, estimation accuracy and trajectory smoothness can be penalized once optimization takes longer. This conclusion is supported by referring to the tightly coupled fusion in Table\,\ref{tab: dus_sovler}. Similarly to the optimizer type, considering a large lag size does not contribute significantly. Moreover, even the incremental smoother with a large lag size frequently violates the desired optimization frequency, resulting in inefficient optimization procedures. 

\begin{table}[!t]
\centering
\caption{Estimation Performance of Seq. DUS using Different Solver Configurations.}
\label{tab: dus_sovler}
\begin{tabular}{l|c|c|c|c}
\hline\hline
\multicolumn{1}{c|}{\textbf{Configuration}} & \textbf{\begin{tabular}[c]{@{}c@{}}Mean 2-D\\ Pos. Err. \\ (m)\end{tabular}} & \textbf{\begin{tabular}[c]{@{}c@{}}Mean 3-D\\ Pos. Err.\\ (m)\end{tabular}} & \textbf{\begin{tabular}[c]{@{}c@{}}Mean \\ Yaw Err.\\ (°)\end{tabular}} & \multicolumn{1}{c}{\textbf{\begin{tabular}[c]{@{}c@{}}Smooth.\\ $s$\end{tabular}}} \\ \hline
lc-batch-3sec                          & 0.324                                                                       & 0.839                                                                      & 0.208                                                                   & 1526.4           \\ \hline
lc-isam2-3sec                          & 0.397                                                                       & 1.146                                                                      & 0.216                                                                   & 5672.8           \\ \hline
tc-batch-3sec                          & 1.543                                                                       & 2.008                                                                      & 0.497                                                                   & 1649.3           \\ \hline
tc-isam2-10sec                         & 1.419                                                                       & 1.733                                                                      & 0.573                                                                   & 1537.9           \\ \hline
tc-isam2-3sec                         & 1.524                                                                       & 1.766                                                                      & 0.511                                                                   & 1510.5           \\ \hline\hline
\end{tabular}
\vspace{1ex}
\raggedright 
Configuration: (fusion type)-(solver type)-(lag size), e.g., lc-batch-3sec
\end{table}

\subsection{GP-WNOA/WNOJ Motion Model}\label{sec: res_wnoj}
In this section, we evaluated the continuous-time trajectory representation using the Gaussian process interpolation with both WNOJ and WNOA models. For a fair evaluation, the hyper-parameter $\myFrameVec{q}{c}{}$ was manually tuned by penalizing the vehicle pose in $\myFrameVec{Q}{c}{}$ with the same parameterization for pose weighting, aiming to evaluate the effect between the third-order and second-order dynamics models.

\begin{figure}[!t]
    \centering
    \includegraphics[width=0.48\textwidth]{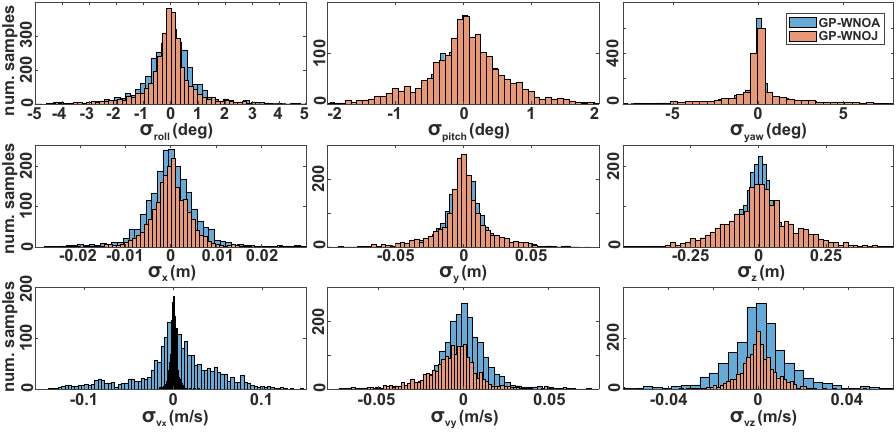}
    \caption{Histogram of whitened state errors of GP models in an open-sky area.}
    \label{fig: gp_hist}
\end{figure}

Compared to the GP-WNOJ model, a GP-WNOA model assumes that the system transition follows a constant velocity model \cite{Barfoot2014BatchCT, fgo_zhang}. As discussed in \cite{WNOJ}, representing vehicle trajectories with an approximately constant-velocity model may be insufficient in urban driving scenarios where the vehicle accelerates and brakes frequently. To evaluate the performance of both GP models, we chose a part of Seq. AC containing $\SI{200}{\s}$ test run in open-sky areas where the PVT solution from the high-grade GNSS receiver presents the ground-truth trajectory. We calculate the whitened error of the vehicle pose and the linear velocity in the body frame and plot the results on the histogram in Fig.\,\ref{fig: gp_hist}. Because the GP-WNOJ model represents second-order system dynamics, it shows smaller errors in all linear velocity components. Both models perform similarly in position estimation, where the GP-WNOJ is more accurate in the main motion direction $x\mathrm{-axis}$. For rotation, the GP-WNOJ does not present considerable improvements compared to the GP-WNOA. One possible reason supporting this result can be traced back to rotational acceleration that cannot be observed directly using the IMU (see Sec.\,\ref{sec: state}). 

We have validated that the GP motion model formulates a valid continuous-time trajectory representation. However, tuning the power spectral matrix $\myFrameVec
{Q}{c}{}$ that scales the system transition in the Gaussian process kernel has a large effect on numerical stability and estimation performance \cite{WNOJ}. Although the GP-WNOJ model presents reliable velocity estimates compared to the GP-WNOA model, it shows higher sensitivity on the power spectral matrix $\myFrameVec
{Q}{c}{}$\cite{WNOJ}, which therefore requires more careful parameter tuning when incorporating accelerations in state propagation.

\begin{remark}
        \textit{\textbf{Tuning of $\myFrameVec{Q}{c}{}$}:} In this work, we did not explicitly investigate parameter tuning for the power spectral matrix $\myFrameVec{Q}{c}{}$. As discussed in the original works \cite{fullsteam, WNOJ}, this hyper-parameter can be calibrated using supervised-learning approaches. Recent work also explored this idea and showed the possibility of learning this parameter without ground-truth labels using variational Bayesian approaches \cite{GP_learning, Wong_2020}. However, vehicle dynamics presents a high variation in real-world driving scenarios, the parameter $\myFrameVec{Q}{c}{}$ should be dynamically tuned in an online process, which remains our future work.
\end{remark}

\section{Conclusion and Future Work}  \label{sec: con}
This article proposes an online factor graph optimization that generalizes multi-sensor fusion for robust trajectory estimation focusing on GNSS. The vehicle trajectory is represented in continuous time using a Gaussian process motion prior that enables arbitrary state querying, presenting a sensor-independent graph optimization. We successfully fused asynchronous sensor measurements into the proposed method for robust vehicle localization in challenging environments. The experimental studies show that the proposed method is robust, flexible, accurate, and works online with multiple datasets collected from challenging scenarios. All our FGO configurations succeed in all test sequences, whereas the classic state-of-the-art lidar-centric method \cite{liosam} failed in some situations. Observed from the experimental results, the GP-WNOJ motion prior enables accurate trajectory representations in continuous time with properly tuned hyper-parameters. 

In this work, we did not fully exploit the GNSS observations, such as carrier-phase, which requires complicated techniques to resolve the ambiguities and detect the satellite cycle slips. Our framework can also utilize advanced techniques to exclude multipath and non-line-of-sight GNSS observations. We also neglected the hyper-parameter tuning of the GP-WNOJ model and sensor noise identification, which can be solved online using learning-based methods. In addition, additional sensor modalities, such as visual odometry, can be utilized in the fusion framework to improve performance. In summary, these research objectives formulate our future work. 

\section*{Acknowledgments}
The authors thank Robin Taborsky from the Institute of Automatic Control at RWTH Aachen University and the public order office in Aachen, Düsseldorf, and Cologne for their great support in measurement campaigns. The authors also thank David Yoon and Keenan Burnett from the Autonomous Space Robotics Laboratory at the University of Toronto for their discussions and support in this work.

\bibliographystyle{IEEETran.bst}
\bibliography{reference}

\begin{IEEEbiography}[{\includegraphics[width=1in,height=1.25in,clip,keepaspectratio]{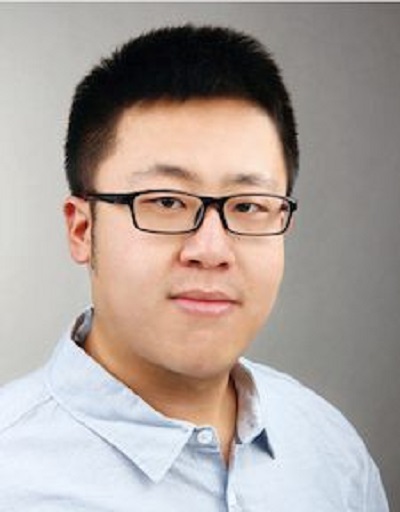}}]{Haoming Zhang} (Member, IEEE) received his MASc degree in mechatronics from the University of Duisburg-Essen, Duisburg, Germany, in 2017. He is now a Ph.D. student at the Institute of Automatic Control at the RWTH Aachen University, Aachen, Germany. His research interests include state estimation, multi-sensor fusion, and learning-based online noise distribution estimation methods. He works on research projects to enable robust vehicle/vessel localization in complex and large-scale environments.
\end{IEEEbiography}

\begin{IEEEbiography}[{\includegraphics[width=1in,height=1.25in,clip,keepaspectratio]{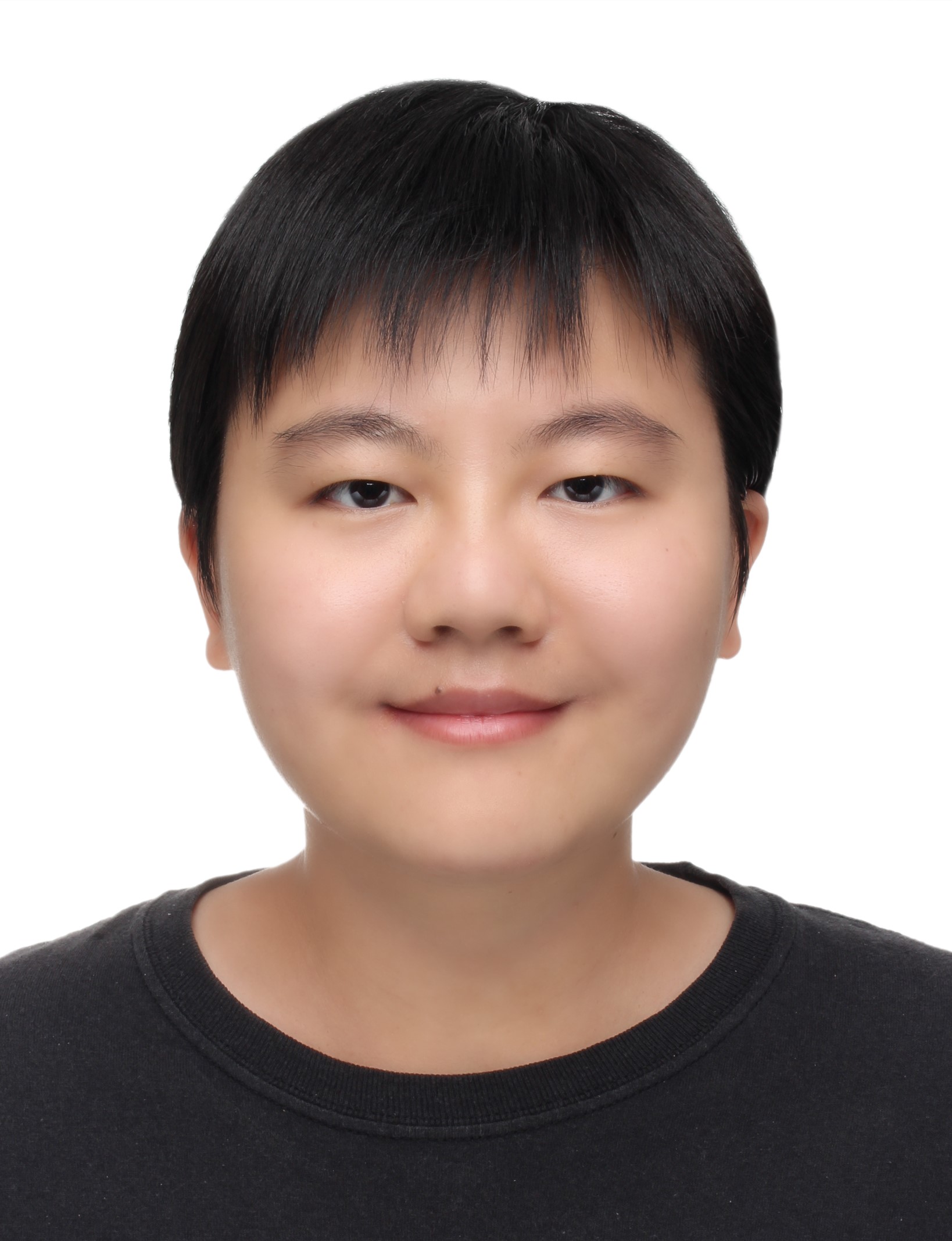}}]{Chih-Chun Chen} (Graduate Student Member, IEEE) received her BASc degree in Engineering Science program (Robotics option) from the University of Toronto, Canada, in 2019 and her MASc degree in Aerospace Science and Engineering at the University of Toronto Institute for Aerospace Studies (UTIAS), in 2021. She is now a Ph.D. student at the Institute of Automatic Control at the RWTH Aachen University, Aachen, Germany. Her research interests include multi-agent path planning, state estimation, and multi-sensor fusion.
\end{IEEEbiography}

\begin{IEEEbiography}[{\includegraphics[width=1in,height=1.25in,clip,keepaspectratio]{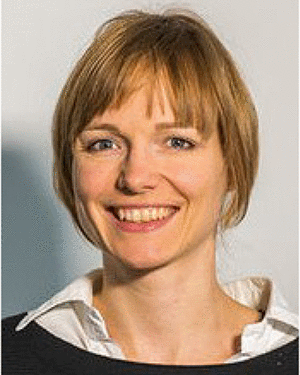}}]{Heike Vallery} (Member, IEEE) graduated from RWTH Aachen University with a Dipl.-Ing. degree in Mechanical Engineering in 2004. In 2009, she earned her Dr.-Ing. from the Technische Universität München and then continued her academic career at ETH Zürich, Khalifa University and TU Delft. Today, she is a full professor at RWTH Aachen and TU Delft, and also holds an honorary professorship at Erasmus MC in Rotterdam. Her main research interests are in design and control of minimalistic robotics.
\end{IEEEbiography}

\begin{IEEEbiography}[{\includegraphics[width=1in,height=1.25in,clip,keepaspectratio]{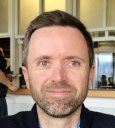}}]{Timothy D. Barfoot} (Fellow, IEEE) received the B.A.Sc. degree in engineering science from University of Toronto, Toronto, ON, Canada, in 1997 and the Ph.D. degree in aerospace science and engineering from University of Toronto, in 2002. He is a Professor with the University of Toronto Robotics Institute, Toronto, ON, Canada. He works in the areas of guidance, navigation, and control of autonomous systems in a variety of applications. He is interested in developing methods to allow robotic systems to operate over long periods of time in large-scale, unstructured, three-dimensional environments, using rich onboard sensing (e.g., cameras and laser rangefinders) and computation. 
\end{IEEEbiography}
\end{document}